\documentclass{article}

\usepackage{arxiv}

\usepackage[utf8]{inputenc} 
\usepackage[T1]{fontenc}    
\usepackage{hyperref}       
\hypersetup{                
    colorlinks=true,        %
    linkcolor=blue,         %
    filecolor=blue,         %
    urlcolor=blue,          %
    citecolor=blue,         %
    }                       %
\usepackage{url}            
\usepackage{booktabs}       
\usepackage{amsfonts}       
\usepackage{amsmath}        
\usepackage{bm}             
\usepackage{leftindex}      
\usepackage{nicefrac}       
\usepackage{microtype}      
\usepackage{fancyhdr}       
\usepackage{graphicx}       
\graphicspath{{media/}}     
\usepackage{subcaption}

\title{Beyond Compression: Training Latent Representations for Stable Long-Horizon Rollout in Neural Surrogate Solvers}

\author{
  Andreas E. Robertson$^{a,b}$ \And
  Ashley T. Lenau$^{a}$\And
  John D. Shimanek$^{a}$\And
  Benjamin A. Jasperson$^{e}$\And
  Vivek Oommen$^{c,f}$\And
  David L. Damm$^{d}$\And
  Krishna Garikipati$^{e}$\And
  Rémi Dingreville$^{a,\star}$\\
  $^{a}$ Center for Integrated Nanotechnologies, Sandia National Laboratories, New Mexico, USA\\
    $^{b}$ Center for Computing Research, Sandia National Laboratories, Albuquerque, NM, USA\\
      $^{c}$ School of Engineering, Brown University, Providence, RI, USA\\
        $^{d}$ Sandia National Laboratories, New Mexico, USA\\
          $^{e}$ Department of Aerospace and Mechanical Engineering, University of Southern California, Los Angeles, CA, USA\\
            $^{f}$ Oak Ridge National Laboratory, Oak Ridge, TN, USA\\
  $^\star$\texttt{\{rdingre\}@sandia.gov} \\
}

\date{September 2026}

\begin{document}

\maketitle

\begin{abstract}
Latent neural surrogate solvers, also called latent dynamics models, seek to accelerate simulations of time-dependent physical systems by recasting their evolution in a compressed latent space rather than resolving full-resolution fields directly. In principle, this reformulation reduces computational cost and can simplify the learning task, in practice, however, errors often accumulate rapidly during long autoregressive rollouts, limiting predictive utility. 
We demonstrate that this instability does not stem from the latent representation itself, instead, it arises when they are trained solely for reconstruction, producing representations poorly suited to long-horizon forecasting.
To address this limitation, we systematically evaluate training-level interventions that align latent representations with long-horizon rollout: Koopman operator learning and Hamming noise injection during autoencoder training to improve compression, together with noise injection and multi-step rollout fine tuning for improving dynamics. 
Importantly, interventions that improve long-horizon rollout stability often degrade conventional training metrics, including reconstruction and one-step prediction accuracy.
Collectively, these training-level interventions reduce long-rollout error by approximately 40\% and match or exceed the accuracy of full-resolution models on two physics benchmarks.
Further, the model is more efficient, requiring 2 orders of magnitude fewer floating point operations and half the GPU memory.
Applied to mesoscale crystal-plasticity simulations of high-cycle fatigue, the resulting surrogate achieves stable extrapolation over horizons orders of magnitude beyond those observed during training.
More broadly, these results show that neural compression should be designed not merely to reduce dimensionality, but to restructure the solution space for stable dynamical evolution, a key for building reliable, efficient neural surrogates for scientific applications.
\end{abstract}

\section{Introduction}\label{sec:intro}
{
Neural surrogate solvers for time-dependent partial differential equations (PDEs) seek solutions orders of magnitude faster than standard numerical methods, in part by advancing for instance the solution over time increments far larger than those resolved by the underlying numerical simulation.
These benefits are especially valuable in scientific applications where high-fidelity simulation is prohibitively expensive.
Among these approaches, latent dynamics models (LDMs)~\cite{montes2021accelerating,yang2021self}, which perform rollout on a compressed (low-dimensional) representation rather than the full-resolution field, are particularly appealing: on standard benchmark problems such as phase-field simulations~\cite{dingreville2024benchmarking} for instance, they can be two to three orders of magnitude faster than comparable pixel-space and direct numerical solvers, using an order of magnitude fewer parameters~\cite{montes2021accelerating,yang2021self, hu2022accelerating, oommen2022learning}.
Additionally, by reducing dimensionality, they open realistic pathways towards adopting memory and compute intensive state-of-the-art architectures (e.g., transformers and diffusion models) \cite{rozet2025lost}. 
Of particular interest are sequentially trained LDMs, where the compression mechanism is optimized first, commonly purely for reconstruction accuracy, and only afterward, a separate dynamics model is trained to predict trajectories within the resulting compressed space.
This simpler strategy has shown to be more scalable to large, expressive architectures and complex problems \cite{eivazi2020deep,linot2020deep,pant2021deep}.
In this paper, we refer exclusively to sequential LDMs.

The benefits of LDMs come at a cost: sequential LDMs are also the least stable over long rollouts, accumulating error more rapidly than their pixel-space counterparts~\cite{dingreville2024benchmarking, oommen2022learning, oommen2024rethinking}.
Such instability is more than a performance gap; it undermines the confidence that practitioners need before integrating these models into larger computational workflows, where extrapolation behavior and the accumulation of error over many iterative model calls determine whether predictions can be trusted at all.
Overcoming this gap is therefore critical not only to realizing the benefits LDMs promise, but to making them viable tools for real scientific and engineering practice~\cite{boyce2023machine}.
}

{
Rollout instability is not incidental to LDMs, but follows from how they are conventionally trained and how the compressed representation at their core is constructed.
These models rest on the manifold hypothesis: the assumption that a system's high-dimensional dynamics evolve on a much lower-dimensional manifold, and that a compression mechanism, most commonly an autoencoder, can be trained to gain access to this manifold~\cite{Fefferman2016testing, desai2024trade}.
Because the first compression training step in sequential training never observes the rollout task, the representation it produces can be misaligned with the requirements of long-horizon prediction~\cite{desai2024trade}, even though it remains highly effective for reconstruction.
Resolving this misalignment is the central objective of this paper.
}

{
The strategy presented in this paper draws on a parallel shift underway in generative modeling and computer vision, where autoencoders are increasingly trained to produce representations tailored to downstream tasks rather than solely to minimize reconstruction error.
For instance, vector quantized generative adversarial networks (VQGANs) encode images into discrete sequences of perceptually meaningful tokens, enabling standard autoregressive transformers to perform image generation~\cite{esser2021taming}.
Similarly, the Genie and Cosmos models tokenize video into discrete latent sequences and then roll out future frames autoregressively within that tokenized space~\cite{bruce2024genie,agarwal2025cosmos}, closely paralleling the latent-space rollout task in LDMs.
Latent diffusion models take a related approach, transferring the diffusion process itself from pixel space to a compressed continuous representation, substantially reducing computational cost while preserving image fidelity~\cite{rombach2022high, sadat2024litevae}.
This perspective is beginning to appear in scientific applications as well: Generale et al.~\cite{generale2024inverse} use a beta-variational autoencoder (VAE) to transform irregular microstructure distributions into standard Gaussian priors for Bayesian inversion in materials design.
Collectively, these examples point to a common lesson: the compression mechanism at the heart of a latent-space model is not fixed to a single objective, but can be trained to emphasize whatever structure the downstream task requires.
A lesson we apply directly to LDMs in this paper.
}

\begin{figure}
    \centering
    \includegraphics[width=0.85\textwidth, height=0.8\textheight, keepaspectratio]{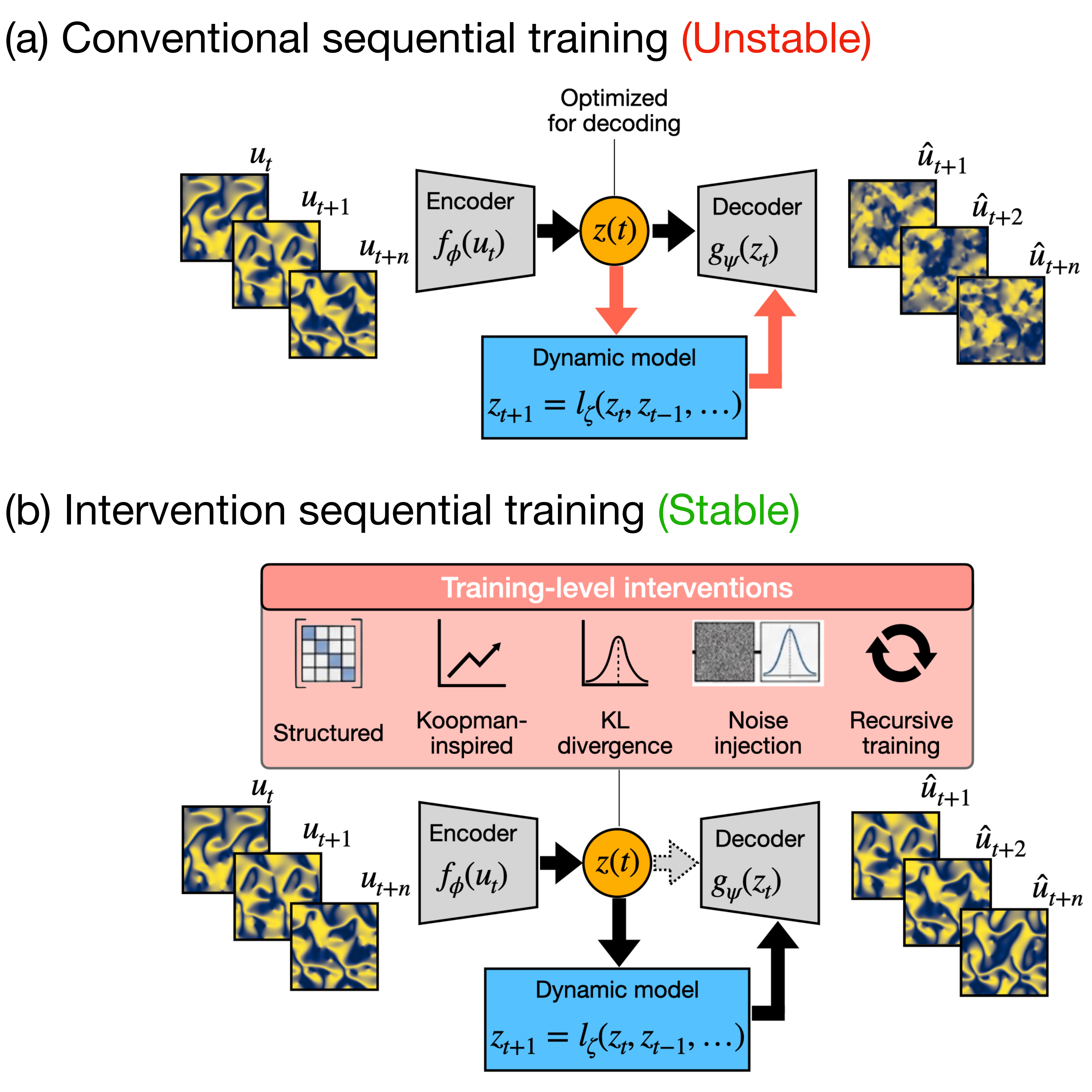}
    \caption{\textbf{Stabilizing latent dynamics models.}
    Comparison between conventional sequential training (where compression and dynamics and trained separately and in sequence) and sequential training with proposed interventions. (a) The autoencoder compression model is trained for reconstruction (black compute path), producing misalignment for long-horizon dynamics predictions (red compute path). 
    (b) Training-level interventions are applied to the autoencoder training leading to better alignment with the dynamics task and stronger long-horizon rollout predictions. Models are still trained sequentially.}
    \label{fig:method}
\end{figure}

{
In this paper, we systematically study five training-level interventions for LDMs (see Figure~\ref{fig:method}):
spatially structured latent compression,
Koopman-inspired constraints on the (latent) dynamics,
Kullback-Leibler (KL) divergence regularization,
noise injection during training, and
recursive training on stabilizing performance.
Rather than pursuing architectural optimization, we isolate the effect of these training and framework decisions directly.
We study three dissimilar dynamic evolution problems -- spinodal decomposition, active matter evolution, and crystal-plasticity fatigue simulations -- each of which poses a distinct challenge for latent dynamics prediction:
sharp interface morphological evolution;
fast, high-dimensional, co-evolving fields; and
dynamics that must be extrapolated many orders of magnitude beyond the training horizon.
Guided by long-horizon performance rather than single-step accuracy, we uncover an important pattern: beneficial interventions often degrade one-step predictive accuracy while substantially improving long rollout predictions (Section~\ref{sec:accumulated-improvement}).
This is a direct result of misalignment between standard training tasks and the real objective of these neural solvers -- stable, long rollouts.
Altogether, through these adjustments, we achieve a $40\%$ reduction in long rollout errors for the active matter problem, for example.
In addition, we see that state-of-the-art performance is achievable using 2 orders of magnitude less compute and half the GPU memory.
Further, we demonstrate these gains in a neural surrogate model for high-cycle fatigue prediction in mesoscale crystal-plasticity simulations that remain stable far beyond the training horizon, using only computationally accessible simulation data as input.
}

\section{Related work}\label{sec:relwork}
{
This work sits at the intersection of three related topics:
the use of latent-space representations to accelerate neural surrogates for time-dependent physical systems,
ongoing efforts to diagnose and mitigate rollout instability in autoregressive neural solvers, and
a growing recognition, first observed in computer vision, that autoencoders can be trained to support a downstream task rather than simply compress their input.
We review each topic in turn below.
}

{
Latent-space models have been widely adopted in scientific machine learning as a means of accelerating neural surrogate solvers for large, time-dependent physical systems~\cite{montes2021accelerating,yang2021self,oommen2022learning,hu2022accelerating,ahmad2023accelerating,gaikwad2025deep}.
Autoencoder-recurrent architectures, for instance, have been used to accelerate phase-field and microstructure-evolution simulations, coupling a convolutional autoencoder with a recurrent dynamics model to propagate the compressed state forward in time~\cite{hu2022accelerating,ahmad2023accelerating}.
More recently, the `Universal Physics Transformers' proposed by Alkin et al.~\cite{alkin2024universal} modernized this idea within a transformer-based framework, propagating dynamics through scalable latent representations that accommodate large, heterogeneous meshes and particle systems.
Across these efforts, however, the autoencoder is most commonly trained purely to reconstruct its input, independent of the rollout task the resulting representation will ultimately support.
Dingreville et al.~\cite{dingreville2024benchmarking} directly compared this sequentially trained latent dynamics approach against a pixel-space solver on a common phase-field benchmark and found a substantial performance gap: despite its far greater computational efficiency, the LDM accumulated rollout error more rapidly and less predictably than its pixel-space counterpart.
Oommen et al.~\cite{oommen2022learning} reported a related sensitivity to the quality of the learned latent representation in similar microstructure-evolution settings.
This body of work highlights the specific gap the present study addresses: unlike the computer-vision efforts discussed in the Introduction, which increasingly train autoencoders around a downstream objective, scientific applications of LDMs have so far continued to optimize the autoencoder purely for reconstruction, leaving this performance gap unresolved.
}

{
In parallel, substantial effort has focused on diagnosing the failure modes of autoregressive neural solvers more broadly, largely independent of whether a compressed representation is used at all.
Recent analyses have connected rollout instability to spurious spectral content introduced by pixel-space regression objectives~\cite{worrall2025spectral} and to grid-aligned errors introduced by patch-based transformer tokenization~\cite{mccabe2025walrus}.
Both argue that these failure modes drive predictions off the training manifold, producing compounding error over successive rollout steps.
A number of empirical mitigation methods have been proposed in response.
For instance, models like the MeshGraphNets counteract this distribution shift by injecting noise into the training data~\cite{pfaff2020learning}, while FourCastNet incorporates multistep training to expose the model to its own prediction errors during optimization~\cite{pathak2022fourcastnet}.
PDE-Refiner instead performs iterative denoising to recover spatial-frequency components that are poorly represented by conventional regression objectives~\cite{lippe2023pde}.
Notably, two of these strategies, namely noise injection and multistep training, are central to the training-level interventions we study in this paper.
However, prior work has applied them directly to the dynamics model operating in pixel space, whereas we apply them to the training of the compression mechanism itself, to shape the latent representation rather than only correct the dynamics model's predictions downstream.
}

{
Finally, a small number of recent efforts have begun applying task-aware representation learning directly to scientific machine learning, using latent representations not only to accelerate rollout but to improve its transferability and predictive stability.
One such model, TadPole, pretrains an autoencoder as a foundation model for three-dimensional PDE fields, reintroducing encoder–decoder skip connections during fine-tuning and rollout to recover a pixel-space UNet architecture \cite{liu2026tadpole}.
Similarly, Rozet et al.~\cite{rozet2025lost} adopt a latent diffusion framework for rollout prediction and show that the resulting diffusion predictor can incorporate partial observations through guided sampling, using this mechanism to stabilize long-horizon predictions.
The present work pursues the same broader objective but targets sequentially trained LDMs directly: rather than introducing new architectural pathways between encoder and decoder, or replacing the dynamics model with a diffusion process, we modify how the autoencoder itself is trained, improving the representation's suitability for autoregressive rollout while preserving the efficiency and implementation simplicity of separately trained autoencoding and dynamics models.
}

\section{Methods}\label{sec:methods}

\subsection{Baseline latent dynamics model formulation}\label{sec:baseline-ldm}

{
A LDM consists of two components trained sequentially:
(i) an autoencoder, comprised of an encoder, $f_\phi$, and a decoder, $g_\psi$, which compresses an ambient-space field $u_t$ into a latent encoding $z_t = f_\phi(u_t)$ and reconstructs it as $\hat{u}_t = g_\psi(z_t)$ respectively.
And (ii) a dynamics model, $l_\zeta$, which predicts the evolution of the latent encoding forward in time, $z_{t+1} = l_\zeta(z_t, z_{t-1}, \dots, P)$, where $P$ denotes the governing partial differential equation (PDE) parameters.
The dynamics model is given a truncated history, $z_t, z_{t-1}, ...$ to inform its predictions.
Composing all three networks gives a full autoregressive update operator,
\begin{equation}
    D_\theta(u_t, u_{t-1}, \dots) = g_\psi\Big( l_\zeta\big( f_\phi(u_t), f_\phi(u_{t-1}), \dots, P \big) \Big).
    \label{eq:operator}
\end{equation}
In practice, the dynamics model is applied repeatedly in latent space to advance many time steps before the decoder is invoked to return to the ambient space.
Following standard practice, the autoencoder and dynamics model are trained sequentially: the autoencoder is optimized first, purely for reconstruction accuracy, and only afterward the dynamics model is trained on the resulting latent trajectories, with the autoencoder held fixed.
}

{
As discussed in the Introduction, this paper is focused primarily on training strategy.
As a consequence, we forgo extensive architecture optimization and, instead, adopt current state-of-the-art architectures:
an axial-vision transformer (AViT)~\cite{mccabe2024multiple, ho2019axial, mccabe2025walrus}, used as the dynamics model $l_\zeta$;
and a wavelet autoencoder~\cite{sadat2024litevae}, used as the compression mechanism $f_\phi, g_\psi$.
Full architectural specifications, loss terms, and optimization for all models are provided in Appendix~\ref{apdx:architecture}.
Additionally, when informative, we adopt standard baselines for each case study:
a UNet~\cite{dingreville2024benchmarking, ho2020denoising}, used as a pixel-space baseline for spinodal decomposition and a standard LDM (using a VQVAE~\cite{esser2021taming}-type autoencoder with additional dense layers for vectorization and a standard transformer as the latent dynamics model) for the active matter problem.
}

\subsection{Training-level interventions for stabilizing rollout}
\label{sec:interventions}
{
We study the following interventions applied to the LDM described above.
}

{
{\bf Spatially structured latent encodings:} The purpose of this intervention is to test whether the degree and structure of compression itself, independent of any other training choice, affects how well the latent representation supports long-horizon rollout.
The original LDM considered in this work~\cite{dingreville2024benchmarking} compresses each field into a flat vector of latent values, produced by a convolutional autoencoder with a small, fully connected bottleneck at its center.
This design favors aggressive compression, reaching ratios of up to $500{:}1$ between the number of input pixels and the size of the latent encoding.
Here, we adopt a fully convolutional wavelet-based autoencoder~\cite{sadat2024litevae} (Section~\ref{sec:baseline-ldm}) whose latent encoding retains spatial structure, effectively a smaller image rather than a flat vector, and compresses far less aggressively, at a ratio of just $64{:}1$.
Because this latent encoding retains spatial structure, we replace the baseline's transformer-based dynamics model with the AViT architecture (Appendix~\ref{apdx:architecture}), which is designed to operate directly on image-like latent representations.
This is the only architectural intervention considered.
}

{
{\bf Koopman-inspired dynamics constraints:} Koopman operator learning is a strategy for regularizing the latent space by promoting an approximately linear latent dynamical system during autoencoder training~\cite{geneva2022transformers}.
Specifically, Koopman learning frameworks concurrently optimize the encoder/decoder and a matrix operator $K$ such that subsequent latent codes are linearly related: $z_{t+1} = Kz_t$.
Unlike standard Koopman learning \cite{lusch2018koopman}, in Koopman operator learning $K$ is discarded after autoencoder training and replaced by expressive deep learning model; only the resulting autoencoder, $f_\phi$ and $g_\psi$, not the linear operator itself, is retained for the second-stage dynamics model.
Following this approach, we add a second reconstruction term to the autoencoder loss that penalizes the reconstruction of the subsequent time step, $\hat{u}_{t+1} = g_\psi\big(K(z_t)\big), z_t = f_\phi(u_t)$, where $K$ is a learned Koopman operator, implemented here as a bias-free, kernel-$3$ convolution rather than the banded linear transformation used in prior work, to accommodate the image-structured latent space.
An additional penalty, $\|K\|$, on the magnitude of the convolution kernel prevents arbitrary rescaling of the latent space~\cite{geneva2022transformers}.
Unlike standard Koopman learning, the Koopman reconstruction loss term is only weakly penalized ($1:2$ with the primary reconstruction loss) to encourage linear dynamics, but not at the expense of degradation in reconstruction.
}

{
{\bf KL divergence regularization:} Because the latent space is otherwise free to expand arbitrarily, for this intervention, we consider a KL divergence penalty encouraging the latent encoding to adopt a unit Gaussian structure, a standard strategy for constraining latent space growth~\cite{esser2021taming, sadat2024litevae}.
This term is defined as $\mathcal{L}_{KL} = \left\langle \left(\langle z_t \rangle_b\right)^2 \right\rangle_c + \left\langle \left( \left\langle (z_t - \langle z_t \rangle_b)^2 \right\rangle_b - 1.0\right)^2 \right\rangle_c$, where $\langle \cdot \rangle_b$ denotes the mean over the batch and $\langle \cdot \rangle_c$ denotes the mean over all remaining components (e.g., spatial and channel dimensions of the latent encoding).
}

{
{\bf Noise injection:} Noise injection trains the model to correct potential errors from encoding or rollout.
We consider noise injection at two stages in the LDM training.
First, we apply \emph{latent-space noise injection} (we call this Hamming noise injection): following the encoding step and prior to the Koopman transformation, we inject constant-variance white Gaussian noise directly into the latent encoding.
This noise forces the autoencoder to separate latent points, learning representations that are robust to corruption (analogous to the Hamming ball in Hamming codes~\cite{hamming1950error}).
This latent structure simplifies the dynamics task: 
it separates nearby latent points so that the dynamics model only needs to target a small region around each encoded point rather than the exact point.
Second, we apply \emph{dynamics-side noise injection}: independent of the autoencoder, we inject white noise into the AViT's input during dynamics-model pretraining, exposing the dynamics model itself to perturbed latent trajectories~\cite{pfaff2020learning}.
}

{
{\bf Recursive (multistep) training:} Finally, we consider recursive training of the dynamics model, in which the AViT is fine-tuned after pretraining by rolling out its own predictions for a fixed number of additional steps before the loss is computed only on the final output, following the multistep strategy used by Pathak et al.~\cite{pathak2022fourcastnet} and their FourCastNet model.
This exposes the dynamics model to its own accumulated prediction errors during training, forcing it to learn to correct for its own mistakes rather than assuming access to ground-truth latent context at every step.
We explore fine-tuning with up to three additional rollout steps.
}

\subsection{Performance and diagnostic metrics}\label{sec:metrics}
{
We use two types of metrics throughout this paper:
performance metrics (relative mean squared error and variance-scaled root mean squared error), which quantify prediction accuracy over rollout, and
two diagnostic metrics (decoder sensitivity and scaled latent space velocity), which help explain why a given intervention improves or degrades that performance.
}

\subsubsection{Performance metrics}
{
For a predicted field $\hat{u}$ and target field $u$, we report the relative mean squared error (RelMSE) and the variance-scaled root mean squared error (VRMSE) as:
\begin{equation}
    \mathrm{RelMSE}(\hat{u}, u) = \frac{\left \langle (\hat{u} - u )^2 \right \rangle}{\left \langle u^2\right \rangle},\quad \mathrm{VRMSE}(\hat{u},u) =  \sqrt{
    \frac{\left\langle(\hat{u}-u)^2\right\rangle}
    {\left\langle\left(u-\langle u\rangle\right)^2\right\rangle}
    },
\end{equation}
\noindent where $\langle\cdot\rangle$ denotes spatial averaging.
Both metrics are computed on a per-time step basis.
VRMSE has a direct interpretive baseline.
A value of VRMSE=0 indicates perfect prediction, while a value of VRMSE=1 corresponds to the error obtained by predicting the spatial mean of the target; values above VRMSE=1 therefore perform worse than this mean-field baseline.
}

\subsubsection{Diagnostic metrics}
{
We utilize two diagnostic metrics to analyze the quality of the trained autoencoders.
The first metric, variance-scaled latent velocity, $v$, is closely inspired by Desai et al.~\cite{desai2024trade} and aims to quantify the complexity of the latent time series.

\begin{equation}
    v(z_t, z_{t+1}) = \| \frac{z_{t+1} - z_t}{\Delta t} \| / \sqrt{\sigma_z^2}.\label{eq:velocity}
\end{equation}

Here, $\Delta t$ was set to 1 arbitrarily and $\sigma_z^2$ is the variance of the latent encoding.
This metric measures the tortuosity of the latent encoding $z$.
The variance scaling normalizes the calculation to the length scale of the learned latent space.

The second metric, decoder sensitivity, measures the sensitivity of the reconstruction to perturbations in the latent space.
This quantifies the extent to which errors made in the latent space will be translated into the real space. 
%
%
We quantify this sensitivity for a decoder $f_\theta: \mathbb{R}^{d_z} \to \mathbb{R}^{d_o}$ with latent encoding $z \in \mathbb{R}^{d_z}$ as the coordinate-wise root-mean-square sensitivity,
\begin{equation}
    s_j(z) = \sqrt{ \frac{1}{d_o} \sum_{i=1}^{d_o} \left( \frac{\partial f_{\theta,i}(z)}{\partial z_j} \right)^2 } = \sqrt{ \frac{1}{d_o} \left[ J(z)^\top J(z) \right]_{jj} },
    \label{eq:sensitivity-jacobian}
\end{equation}
with the decoder Jacobian $J(z) = \partial f_\theta(z) / \partial z$.
Large values of $s_j(z)$ indicate that small perturbations in latent coordinate $j$ produce large changes in the decoded output.
Conversely, small values indicate that the decoder is locally insensitive to that coordinate, which suppresses the propagation of latent-space errors, an especially valuable property during rollout.
Because explicitly forming $J(z)$ is computationally prohibitive for the high-dimensional outputs considered in this paper, we estimate $s_j(z)$ using a Hutchinson-style randomized estimator~\cite{hutchinson1989stochastic,bekas2007estimator}.
The full derivation is provided in Appendix~\ref{apdx:sensitivity}.
}

\subsection{Foundation model: Walrus}\label{sec:walrus-foundation-model}
One of the envisioned benefits of the spatially structured latent encodings introduced in Section~\ref{sec:interventions} is that they may be compatible with modern physical foundation models pretrained directly on real-space data.
Specifically, features learned during a foundation model's physical pretraining in real space are likely to remain relevant when applied within the compressed latent space of an LDM, provided that space retains enough physical structure.

Motivated by this, in the Discussion Section, we will briefly explore the applicability of the Walrus foundation model~\cite{mccabe2024multiple,mccabe2025walrus}, a 1.3-billion parameter transformer-based foundation model pretrained across nineteen physical scenarios spanning fluid dynamics, acoustics, and astrophysics, as a latent dynamics surrogate.
The model employs a space-time factorized architecture that alternates attention along the spatial and causal temporal axes to capture complex multi-dimensional field dynamics. 
Here, the Walrus model serves as an alternative dynamics model $l_\zeta$, in place of the AViT described above. 
The aim of this small secondary experiment is to explore whether any potential alignment can be leveraged to accelerate the training of the LDM.


%

\section{Datasets and case studies}\label{sec:datasets}
{
We evaluate the training-level interventions described above across three case studies of dynamic physical evolution, each selected to stress a different aspect of latent dynamics prediction:
a single-field, sharp interface evolving morphology (spinodal decomposition);
several fast, tensor-coupled fields with an inherent closure problem (active matter); and a high-dimensional, multi-physics system that must be extrapolated far beyond its training horizon (high-cycle fatigue).
Full simulation parameters and data generation procedures for all three case studies are provided in Appendix~\ref{apdx:dataset}.
}

\subsection{Spinodal decomposition}\label{sec:spinodal-casestudy}
{
Spinodal decomposition describes the phase separation of an initially homogeneous mixture into two coexisting phases, modeled using the Cahn--Hilliard equation: a single conserved concentration field evolving under a fourth-order gradient-flow partial differential equation~\cite{dingreville2024benchmarking}.
The dataset is comprised of 1\,000 simulations generated by Latin hypercube sampling over four physical parameters (average concentration, gradient-energy coefficient, mobility, and barrier height), each tracked on a $384 \times 384$ grid over 200 recorded time steps.
This case study serves primarily as a controlled point of comparison to the literature: its single field and comparatively slow dynamics, outside of the initial transient, let us verify that our training interventions recover performance previously attributed only to pixel-space models.
Full dataset details and governing equations are provided in Appendix~\ref{apdx:dataset}.
}
\subsection{Active matter}\label{sec:activematter-casestudy}
{
The active matter case study, drawn from the Well benchmark suite~\cite{ohana2024well}, models a dense suspension of active particles in an incompressible Stokes fluid, coupling a Smoluchowski equation for the particle orientation distribution to the fluid's momentum balance.
The dataset comprises 225 independent trajectories, each tracked on a $256 \times 256$ grid over 81 recorded time steps, resolving eleven simultaneously coupled fields, including particle concentration, fluid velocity, and orientation and strain-rate tensors, and retains the same fourth-order mathematical structure as the phase-field problem above.
Because the evolution of the resolved orientational moments depends on an unresolved higher-order moment, the problem also contains an implicit closure problem, compounding the challenge posed by its fast dynamics and high dimensionality.
These properties make this case study a substantially more demanding stress test than spinodal decomposition, and we adopt it as our primary testbed for systematically studying each training intervention introduced in Section~\ref{sec:interventions}.
Additionally, because this dataset is drawn from a public benchmark, it lets us directly compare model performance against alternative modeling strategies (although, we emphasize that maximizing performance is not the goal of this paper).
Full dataset details and governing equations are provided in Appendix~\ref{apdx:dataset}.
}
\subsection{High-cycle fatigue}\label{sec:fatigue-casestudy}
{
Our final case study applies the resulting model to even higher dimensional problem: predicting high-cycle fatigue damage accumulation in a polycrystalline metal from crystal-plasticity fast Fourier transform (CPFFT) simulations~\cite{lebensohn2020spectral,dingreville2010effect}.
This problem is particularly expensive because it suffers from two competing time-scales; the phenomena of interest (fatigue damage accumulation) occurs on exponential time scales while conventional numerical solvers operate on linear (often sub-cycle) time scales to retain stability. 
Neural solvers offer a potential solution: leveraging their pattern learning flexibility, they can be trained directly on the problem's natural exponential time scale.
However, even this solution comes with a challenge: generating training data at cycle counts of practical interest is prohibitively expensive; the dataset used here is therefore logarithmically subsampled and limited to simulations of $2^9$ cycles or fewer (i.e., the time series for training includes $\{x(t_{2^0}), x(t_{2^1}), x(t_{2^2}), ..., x(t_{2^9})\}$.
The dataset resolves 55 co-evolving field quantities on a $128 \times 128$ grid.
This case study tests a different capability than the two benchmarks above: rather than stress-testing rollout stability over a fixed horizon, it tests whether a model trained only on computationally accessible cycle counts can extrapolate reliably to cycle counts far beyond its training data, using a relative-time encoding that allows rollout at cycle counts outside the training distribution.
It further tests scalability to a substantially higher-dimensional system, with 55 co-evolving field quantities, a significant increase over both other case studies.
Full details of the microstructure extraction, simulation procedure, and dataset construction are provided in Appendix~\ref{apdx:dataset}.
}

\section{Results}\label{sec:results}
{
This section is organized around three questions designed to probe rollout performance from different angles, rather than around the specific case studies used to answer each one.
Section~\ref{sec:res-baseline} asks whether the training-level interventions introduced in Section~\ref{sec:interventions} are sufficient to close the previously documented gap between latent-space and pixel-space rollout performance.
Section~\ref{sec:res-intervention} isolates the effect of each individual intervention, and examines how these effects accumulate when combined.
Section~\ref{sec:res-fatigue} asks whether the resulting model can extrapolate reliably far beyond its training horizon, a capability distinct from rollout stability over a fixed number of steps.
Throughout, we report performance using the metrics defined in Section~\ref{sec:metrics} applied to the case studies described in Section~\ref{sec:datasets}.
}

\subsection{Performance gaps and baseline improvements validation}\label{sec:res-baseline}

\begin{figure}
    \centering
    \includegraphics[width=\textwidth, height=0.69\textheight, keepaspectratio]{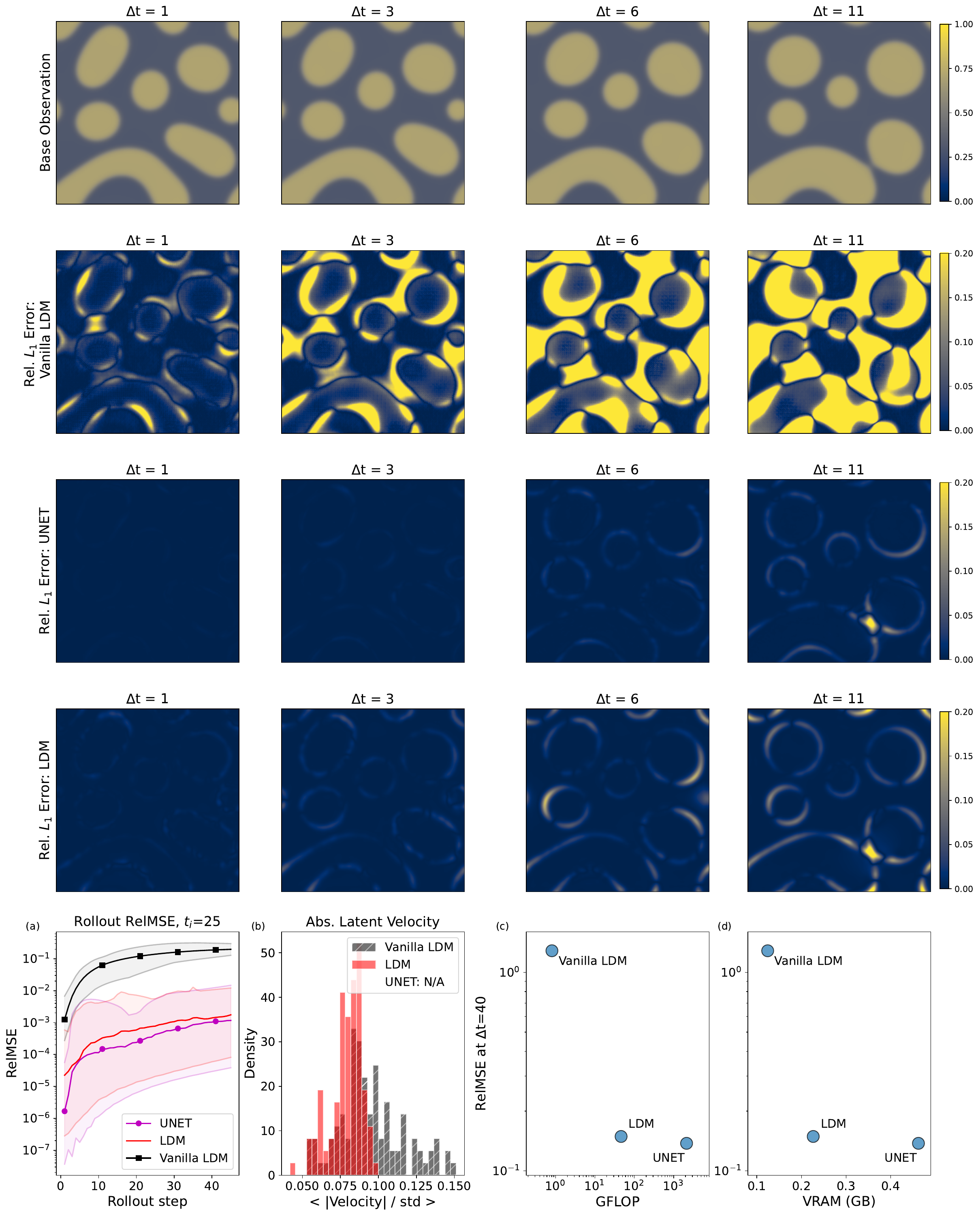}
    \caption{\textbf{Comparison between the pixel-space model dynamic model (UNet) and the proposed LDM incorporating the interventions of Section~\ref{sec:interventions} (LDM) on the spinodal decomposition benchmark.}
    Rows 1--4 show the ground-truth concentration field and the relative $L_1$ error for the vanilla LDM, the UNet, and the proposed LDM, respectively, at rollout steps $\Delta t = 1, 3, 6, 11$; each model is given five initial time steps as context before predictions are generated autoregressively.
    (a) RelMSE as a function of rollout step, averaged across the test dataset, starting at time, $t_i = 25$. 
    (b) Distribution of mean absolute latent-space velocity magnitude for the vanilla LDM and the proposed LDM. The UNET is not applicable to this analysis.
    (c) Model comparison of (Giga) Floating Point Operations (GFLOP) versus RelMSE at $40$ step rollout (averaged over the entire dataset). 
    (d) Model comparison of allocated VRAM (Gigabytes - GB) versus RelMSE at $40$ step rollout (averaged over the entire dataset). 
    }
    \label{fig:phasefield_overview}
\end{figure}

{
We begin by answering a specific question: are the training-level interventions introduced in Section~\ref{sec:interventions} sufficient to close the performance gap between latent-space and pixel-space rollout first documented by Dingreivlle et al.~\cite{dingreville2024benchmarking}?
As such, we revisit the spinodal decomposition benchmark problem presented in Section~\ref{sec:spinodal-casestudy}.
Figure~\ref{fig:phasefield_overview} contrasts the rollout performance of three models:
a baseline LDM trained only for reconstruction (denoted as vanilla LDM),
a pixel-space UNet, and
the proposed Koopman-inspired LDM incorporating the interventions described in Section~\ref{sec:interventions}.
The first and second models utilize identical architectures to those reported in Dingreville et al.~\cite{dingreville2024benchmarking}.
The third model utilizes a UNet-based dynamics model (identical architecture) to remain consistent with the pixel-space comparison.
The top row shows the ground-truth concentration field at four rollout steps, $\Delta t = 1, 3, 6, 11$; the three rows beneath it show the corresponding relative $L_1$ error for the vanilla LDM, the UNet, and the proposed LDM (with all interventions) with all the interventions.
Consistent with the original finding, the vanilla LDM accumulates error rapidly, moving off the manifold of realizable solutions almost immediately, while the pixel-space UNet accumulates error more slowly and produces errors that remain physically plausible, with  features that are misaligned with the ground truth (e.g., deformed or mislocated objects) rather than unrealizable ones.
Applying the proposed interventions from Section~\ref{sec:interventions} closes this gap: the resulting Koopman-inspired LDM's rollout error tracks the pixel-space UNet closely across all reported rollout lengths, recovering performance previously attributed only to pixel-space models.
This qualitative pattern is confirmed quantitatively looking at the RelMSE as a function of rollout step (for starting time $t_0 = 25$) in panel (a):
the vanilla LDM's error grows substantially faster and reaches a higher plateau (${\rm RelMSE}\sim 10^{-1}$) than either the UNet or the proposed LDM (${\rm RelMSE}\sim 10^{-3}$), which track closely together throughout the rollout.
Panel (b) in Figure~\ref{fig:phasefield_overview} offer a mechanistic explanation for this improvement: the mean of the latent-space velocity magnitude $v$ (see Eq\@.~\eqref{eq:velocity}) is substantially lower for the proposed model than for the vanilla LDM, indicating that its latent trajectories evolve more slowly and smoothly, consistent with reduced sensitivity to the small errors that accumulate during rollout.
In addition to achieving comparable error, the proposed LDM provides two practical performance improvements compared to the UNET.
As shown in panels (c,d) in Figure~\ref{fig:phasefield_overview}, the LDM achieves the same error while requiring nearly 2 orders of magnitude fewer floating point operations and utilizing half the GPU memory (these values were calculated for a 40-step rollout). These improvements make the proposed LDM a more economical choice for inference, requiring cheaper, less performant hardware to achieve the same performance.
}

\subsection{Systematic ablation of training-level interventions}\label{sec:res-intervention}
{
Having shown that the combined set of interventions can close the performance gap on for this benchmark, we now ask a second question: which of these interventions actually drives that improvement, and by how much?
We answer this by isolating each of the five training-level interventions introduced in Section~\ref{sec:interventions}, applying them sequentially to the baseline latent dynamics model and analyzing their individual contribution to the accumulated effect.
Although the spinodal decomposition case study is a canonical example in the mesoscale materials space because its sharp interface tests a common materials science specific failure mode for neural solvers, its singular spatial field and relatively slow dynamics outside of the initial time steps limits its ability to stress test our proposed framework.
As a result, we conduct this analysis on the Well's active matter case study (Section~\ref{sec:activematter-casestudy}). We also swap from reporting relative mean squared error to reported VRMSE, a standard metric for the Well benchmark dataset. 
}

\begin{figure}
    \centering
    \includegraphics[width=\textwidth, height=0.73\textheight, keepaspectratio]{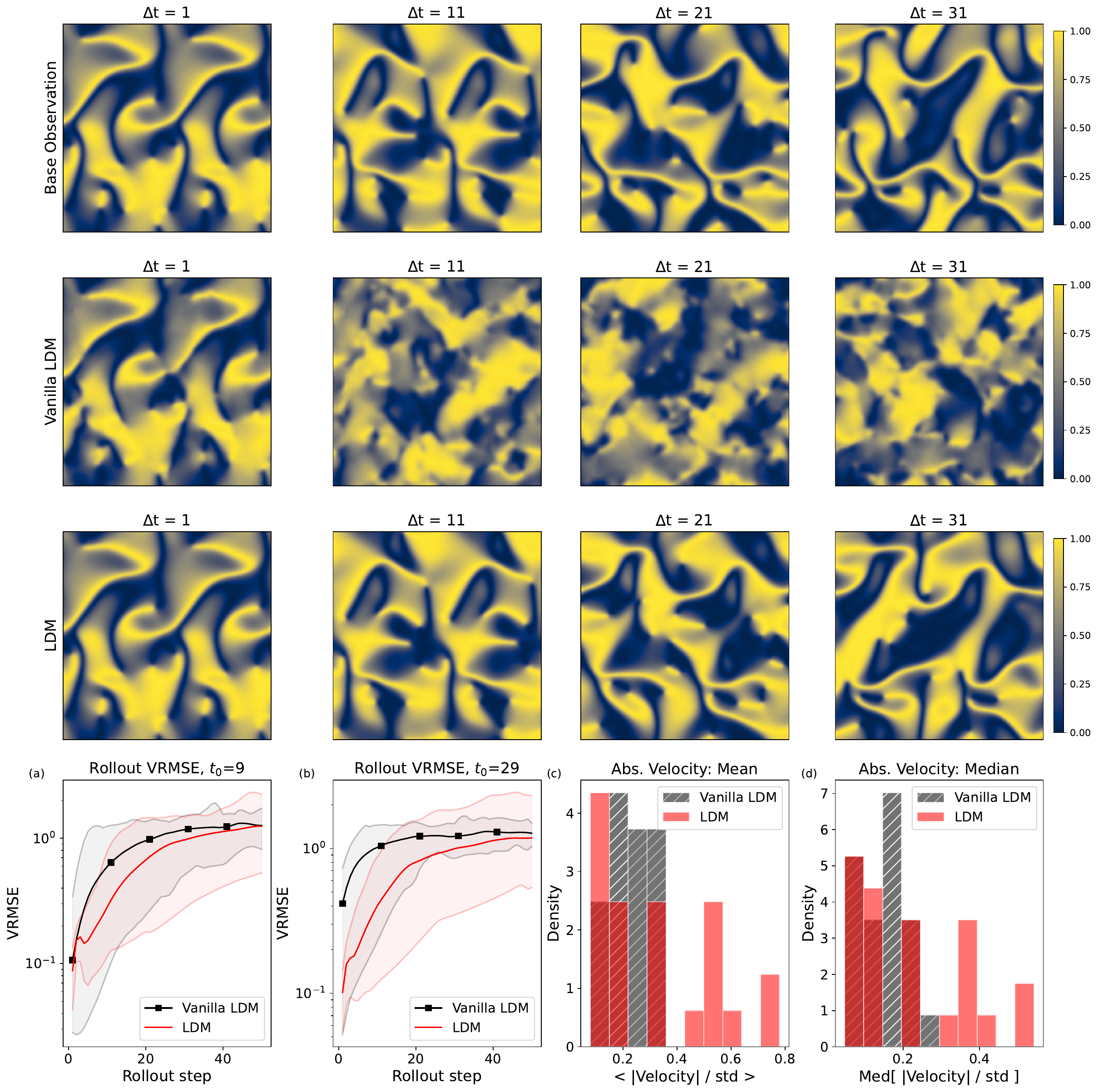}
    \caption{\textbf{Comparison between the vanilla LDM and proposed LDM incorporating the interventions of Section~\ref{sec:interventions} (LDM) on the active matter benchmark.}
    Rows 1--3 show the ground-truth of the first component of the second-order moment tensor, $D_{0,0}$, respectively, at rollout steps $\Delta t = 1, 11, 21, 31$; each model is given $9$ time steps of historical context before predictions are generated autoregressively.
    (a,b) VRMSE as a function of rollout step, averaged across the test dataset, from two starting times, $t_0 = 9$ and $t_0 = 29$. 
    (c,d) Distribution of mean and median absolute latent-space velocity magnitude for the vanilla LDM and the proposed model.}
    \label{fig:activematter_overview}
\end{figure}

Figure~\ref{fig:activematter_overview} previews this section's overall result before we turn to the individual interventions that produce it.
%
%
Here, we make two small changes to the vanilla LDM to isolate performance differences to the proposed interventions.
First, we replace the LSTM dynamics model with a transformer architecture to minimize the confounding impact on our experiments of using a transformer based dynamics model in the new proposed LDM.
Second, we adopt the same training recipe (e.g., learning rate and number of optimization steps) as the proposed LDM.
Figure~\ref{fig:activematter_overview} reveals many of the same performance trends discussed in the spinodal decomposition task previously.
The proposed LDM retains stable predictions with qualitatively good physical alignment up to $21$.
Even at $31$ steps, the feature-types remain consistent, although exact features begin to deviate significantly.
A notable deviation between the two examples is that the velocity profiles of the two latent spaces are much more similar, Figure~\ref{fig:activematter_overview}c,d.
Although differences in the profiles are present, this metric alone does not sufficiently explain the difference in performance.

\subsubsection{Effect of spatially structured latent encodings}
\begin{figure}
    \centering
    \includegraphics[width=\textwidth, height=0.67\textheight, keepaspectratio]{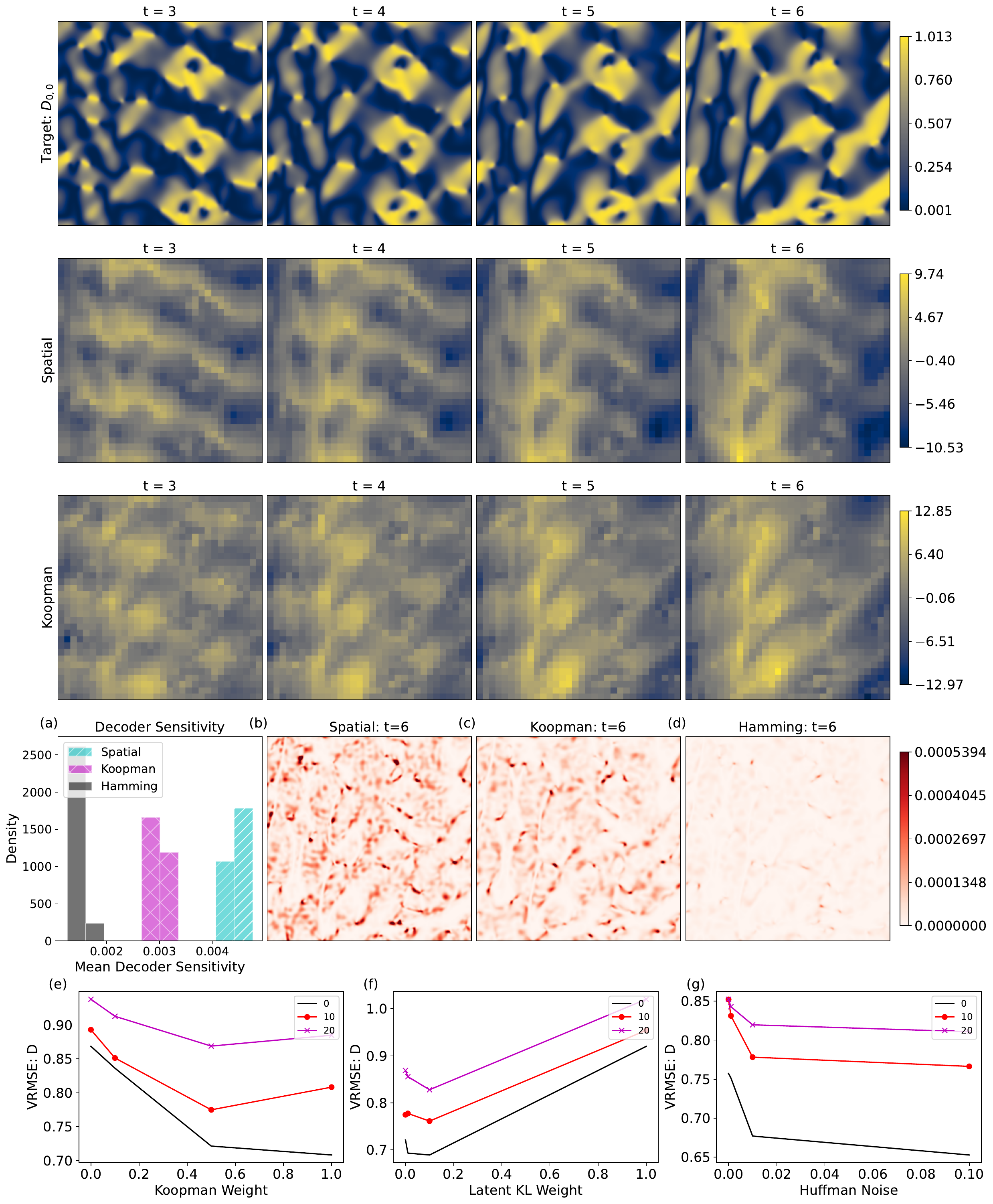}
    \caption{\textbf{Overview of the impact of the spatially structured latent encodings, Koopman-inspired dynamics constraints, and Hamming noise injection interventions.}
    Row 1 depicts real space samples of the first component of the second order moment tensor, $D_{0,0}$.
    Row 2 depicts corresponding samples of one of the $8$ latent encodings from the LiteVAE-type autoencoder.
    Row 3 similarly depicts a single latent field from the Koopman-trained LiteVAE autoencoder.
    (a) Comparison of the decoder sensitivity, Section~\ref{sec:metrics}, between the basic LiteVAE autoencoder, koopman training, and koopman + hamming noise injection.
    (b-d) variance of the reconstruction for the $t=6$ sample depicted in Row 1. Variance was calculated over $30$ samples generated by injecting standard deviation $0.01$ noise onto the latent encoding.
    (e-g) hyperparameter optimization for the hyperparameters involved in proposed autoencoder interventions. Error is reported for the second order moment tensor.
    See Appendix~\ref{apdx:full_parameter_optimization} for optimization over remaining parameters.}
    \label{fig:koopman_latent_code}
\end{figure}

{
The first intervention is adopting the computer-vision standard format for autoencoders: fully convolutional autoencoders which result in spatially structured latent encodings, see Figure~\ref{fig:koopman_latent_code}.
We re-emphasize that our primary focus is on training optimization over architecture optimization.
Therefore, we forgo architecture tuning to the extent possible, preferring to adopt state-of-the-art architectures.
Specifically, we use a LiteVAE-type wavelet autoencoder \cite{sadat2024litevae, esser2021taming}, which are design for more data- and compute-efficient training.
See Appendix~\ref{apdx:architecture} for further details on the architecture.
Further, we use a compression ratio of just $64{:}1$, reduced from $500{:}1$ in the vanilla LDM.
These changes produce latent encoding which is itself an image, displaying more interpretable, physical structure, see, e.g., Figure~\ref{fig:koopman_latent_code}, second row.
Qualitatively, the latent encodings contain features that track the evolution of the primary, real space features.
This correlation points to good dynamic structure in the latent encoding, not a direct correspondence; the real space depicts one of $11$ total fields, which are nonlinearly encoded and compressed into one of $8$ latent fields.
To accommodate the image-structured latent space, we replace the vanilla LDM's transformer-based dynamics model with an axial-in-time, dense-in-space vision transformer (AViT) \cite{mccabe2024multiple, ho2019axial}, see Appendix~\ref{apdx:architecture} for further details.
We evaluate these two changes jointly rather than in isolation, since the switch to an image-structured latent space is what makes the AViT architecture viable in the first place; Sections~\ref{sec:koopman-results}--\ref{sec:recursive-training-results} then isolate the remaining interventions individually against this joint baseline.
}

{
These changes led to an immediate and striking improvement in predictive error for short rollouts:
VRMSE at a rollout of $1$ and $5$ improves by $82.7\%$ and $54.6\%$ respectively ($0.281$ to $0.048$ and $0.593$ to $0.269$, respectively), see Table~\ref{tab:vrmse_by_dt}.
At $15$-step rollout, however, this trend reverses over longer rollouts: performance instead \textit{decreases} by $3.8\%$ relative to the vanilla LDM.
This is the first instance of a pattern that recurs throughout this section:
a change that substantially improves short-horizon accuracy can simultaneously make long-horizon rollout worse, and, as we show in Section~\ref{sec:accumulated-improvement}, the reverse holds just as often.
Appendix~\ref{apdx:add-exp} probes this further, exploring the impact of further optimizing the latent space size (i.e., the spatial resolution and number of latent channels).
This analysis finds clear trends improving reconstruction performance (unsurprisingly: larger latent spaces produce better reconstructions), but these trends do not reliably correlate with long-rollout performance, reinforcing the same lesson:
performance on pretraining and training objectives -- reconstruction (autoencoder) and one-step prediction accuracy (dynamics model) -- is an inconsistent indicator of performance on the primary task of interest, long-rollout accuracy.
From this analysis, we adopt intermediate hyperparameters arbitrarily ($32$ width, or $3$ compression steps, and $8$ latent channels\footnote{Due to a configuration error, some of the subsequent hyperparameter optimization was performed with $11$ channels. This deviation accounts for any difference in reported performance, Appendix~\ref{apdx:add-exp}. This modestly affects the interpretation of the noise injection result in Section~\ref{sec:noise-results}, where we discuss it further.})
Altogether, this intervention increases the flexibility and interpretability of the latent space and provides the architectural foundation for the improvements that follow, but is insufficient on its own to improve long-rollout stability; the following subsections examine whether more targeted training interventions can recover the long-rollout performance lost here.
}

\begin{table}[ht]
\centering
\caption{\textbf{Cumulative effect of training-level interventions on rollout error at different rollout time $\Delta t$.}}
\label{tab:vrmse_by_dt}
\begin{tabular}{|lccc|ccc|ccc|}
\hline
Model & \multicolumn{3}{c|}{$\Delta t=1$} & \multicolumn{3}{c|}{$\Delta t=5$} & \multicolumn{3}{c|}{$\Delta t=15$} \\
\cline{2-10}& Error & Relative & Absolute & Error & Relative & Absolute & Error & Relative & Absolute \\
\hline
Baseline & 0.281 & -- & -- & 0.593 & -- & -- & 0.982 & -- & -- \\
Spatial & 0.049 & 0.827 & 0.827 & 0.269 & 0.546 & 0.546 & 1.020 & -0.039 & -0.039 \\
Koopman & 0.033 & 0.327 & 0.884 & 0.232 & 0.140 & 0.609 & 0.772 & 0.243 & 0.213 \\
KL & 0.036 & -0.100 & 0.872 & 0.241 & -0.038 & 0.595 & 0.906 & -0.174 & 0.077 \\
Hamming Noise & 0.032 & 0.106 & 0.886 & 0.228 & 0.050 & 0.615 & 0.768 & 0.153 & 0.218 \\
Dynamic Noise & 0.044 & -0.381 & 0.842 & 0.192 & 0.158 & 0.676 & 0.601 & 0.217 & 0.388 \\
Roll 1 & 0.063 & -0.427 & 0.775 & 0.180 & 0.063 & 0.696 & 0.624 & -0.039 & 0.364 \\
Roll 2 & 0.087 & -0.373 & 0.691 & 0.170 & 0.055 & 0.713 & 0.579 & 0.073 & 0.410 \\
Roll 3 & 0.094 & -0.081 & 0.666 & 0.182 & -0.072 & 0.692 & 0.558 & 0.036 & 0.432 \\
\hline
\end{tabular}
\end{table}

\subsubsection{Effect of Koopman-inspired dynamics constraints}\label{sec:koopman-results}
{
Adding the Koopman-inspired reconstruction term improves performance across all rollout lengths considered: $32.7\%$, $14.0\%$, and $24.3\%$ improvement at $1$-, $5$-, and $15$-step rollouts, respectively, see Table~\ref{tab:vrmse_by_dt}.
Figure~\ref{fig:koopman_latent_code} (rows 1 and 3) compares the produced latent encoding for an example rollout against the target field.
We emphasize that the latent encoding represents a nonlinear combination of the $11$ co-evolving fields in the ambient space, so a clear one-to-one assignment between a real-space field and a latent channel is rarely possible.
Nonetheless, contrasting the latent encoding without the Koopman loss (Row 2) with the code with the Koopman loss (Row 3) suggests that the Koopman constraint achieves better alignment with the features of the dynamics in latent space than the unconstrained autoencoder.
To test whether the Koopman constraint actually makes the latent dynamics more linear, as Koopman theory would predict, we fit a simple, unconstrained linear model, one not tied to the specific latent structure organization used during training, to predict the next latent state directly from the current one, and measured how well this linear fit explains the data using $R^2$.
Interestingly, the Koopman constraint has an unexpected effect: it actually decreases the linearity of the dynamics.
The fit was slightly \emph{worse} with the Koopman constraint ($R^2 = 0.84$) than without it ($R^2 = 0.92$), and only marginally better than the original baseline model ($R^2 = 0.83$).
Rather than linearizing the dynamics, the Koopman constraint instead produces a less sensitive latent space. 
We measure the sensitivity of the latent space via the decoder sensitivity (see Section~\ref{sec:metrics}).
The Koopman training generates a $~33\%$ improvement in the sensitivity ($0.0045$ to $0.0030$) (the vanilla LDM's baseline sensitivity was $0.0732$), Figure~\ref{fig:koopman_latent_code}a.
Less sensitive decoders will suppress mistakes made in the latent space. This is especially valuable for rollout stability, where errors accumulate aggressively during autoregression, but would be mitigated by a insensitive autoencoder.
Figures~\ref{fig:koopman_latent_code}b,c contrast the variance in reconstruction\footnote{Variance is estimated numerically over $30$ reconstructions produced by adding white noise with standard deviation $0.01$ to the original latent encoding.} between the original spatial autoencoder and one trained with the Koopman loss. The reduced sensitivity leads to lower variance across the field.
}

\subsubsection{Effect of KL divergence regularization}

Because the latent space is otherwise free to expand arbitrarily, see ranges of the latent cencodings in Figure~\ref{fig:koopman_latent_code}, we next consider a KL divergence penalty encouraging the latent encoding toward a unit Gaussion structure to keep growth in check and improve performance on downstream tasks (a standard regularization strategy used particularly generative tasks) \cite{esser2021taming, sadat2024litevae}.
Unlike the two previous interventions, adding this penalty on top of the Koopman constraint lead to inconsistent results producing improvement in predictions for some variables, Figure~\ref{fig:koopman_latent_code}d, and not for others, Figure~\ref{fig:key_parameter_optimization_complete}
As tabulated in Table~\ref{tab:vrmse_by_dt}, relative to the Koopman-only model, adding the KL divergence degrades performance by 10\%, 3.8\%, and 17\% at $\Delta t = 1$, $5$, and $15$, respectively (See Relative column in Table~\ref{tab:vrmse_by_dt}).
Figure~\ref{fig:koopman_latent_code} (f) sweeps the relative weight of the KL term directly and confirms this pattern holds broadly rather than being sensitive to a specific weight. 
We interpret this conflict as a consequence of the two regularizers pulling the latent space in different directions:
the Koopman constraint shapes the latent space specifically around the rollout task,
whereas the KL penalty pushes it toward an isotropic, unit-Gaussian structure irrespective of the underlying dynamics.
Such opposite trends illustrate the broader argument motivating this paper: representation-learning objectives from generative modeling, where KL regularization is standard practice, do not automatically transfer to rollout stability as pursued here. 
Despite this, we adopt a light KL regularization moving forward, loss penalty equal to $0.1$, corresponding to the lowest VRMSE degradation observed.

\subsubsection{Effect of noise injection}\label{sec:noise-results}

{
The biggest improvement in long-rollout error so far was achieved by decreasing the decoder sensitivity. 
We expand on this by injecting noise into the reconstruction process. 
After encoding from the autoencoder, constant-variance white Gaussian noise is added before the convolutional Koopman operator is applied. 
We refer to this process as `Hamming' noise injection, in reference to Hamming codes: its aim is to move latent points apart, effectively producing a small region around each point which the dynamics model can target, instead of a single exact point. 
The goal is to improve the margin for error for the dynamics model's predictions.
}

{
Adding Hamming noise produces a significant improvement in stability realtive to the KL-regularized model: $\sim 20\%$ at $15$-step rollouts, see Table~\ref{tab:vrmse_by_dt}. 
It leads to $\sim 50\%$ reduction in the sensitivity of the latent space, (Figure~\ref{fig:koopman_latent_code}(a)). 
Contrasting Figure~\ref{fig:koopman_latent_code}(c) and (d) shows that adding Hamming noise injection on top of Koopman learning continues to reduce the variance of reconstructions beyond the Koopman constraint alone. 
Hamming  noise injection is beneficial over a large range of noise values tested (we used $0.1$ for final training), see Figure~\ref{fig:koopman_latent_code}g. 
Finally, we observe that Hamming noise injection benefits from the previous KL divergence regularization explored (the $15$-step rollout VRMSE increases $\sim 8\%$ when KL is removed). 
}

{
In addition, following recommendations in MeshGraphNet~\cite{pfaff2020learning}, we explore noise injection into the dynamics model's input during training, Appendix~\ref{apdx:active_matter_avit_ablation}. 
This intervention forces the model to learn to return to the solution manifold when high-frequency perturbations are present. 
Relative to the Hamming-noise model, this dynamics noise injection produces a further $20\%$ improvement in performance at $15$-step rollouts, see Table~\ref{tab:vrmse_by_dt}. 
This benefit is not uniform across rollout lengths, however: at $1$-step rollout, the same intervention degrades performance by $\sim 45\%$, indicating that this parameter is considerably more sensitive than latent-space noise injection, with both too little and too much noise destabilizing training.
}

\subsubsection{Effect of recursive training}\label{sec:recursive-training-results}

Finally, we consider recursive training of the dynamics model:
after pretraining, the AViT is further fine-tuned by rolling out its own predictions for a fixed number of additional steps before the loss is computed only on the final output, following the multistep strategy of FourCastNet~\cite{pathak2022fourcastnet}.
This exposes the dynamics model to its own accumulated prediction errors during training, forcing it to learn to correct for its own mistakes rather than assuming access to ground-truth latent context at every step.
Prior efforts applying this strategy directly in pixel space have attempted to develop memory-efficient approximations to make it tractable~\cite{brandstetter2022message}.
The smaller memory footprint of our compressed latent space lets us perform recursive fine-tuning directly, without approximation, on a single Nvidia A6000 GPU.

Here, we explore up to three additional rollout steps of fine-tuning, see Table~\ref{tab:vrmse_by_dt} (rows \emph{Roll 1--3}).
At $\Delta t = 15$, this reduces raw rollout error from $0.60120$ to $0.55802$, a $7.2\%$ reduction, though the improvement is not strictly monotonic step-by-step.
The first additional rollout step is actually mildly harmful relative to the row above it ($-3.9\%$), while the second rollout step gives us the bulk of the gain ($+7.3\%$), and the third adds a smaller further improvement ($+3.6\%$).
At $\Delta t = 1$, by contrast, all three steps are substantially and consistently harmful ($-42.7\%$, $-37.3\%$, and $-8.1\%$, respectively), degrading much of the short-rollout gain established by the earlier interventions.
Appendix~\ref{apdx:add-exp} finds that the improvement from recursive fine-tuning is observed even under different training recipes for the original autoencoder and the pretrained AViT, indicating that this benefit is not an artifact of a specific training configuration.

\subsubsection{Cumulated improvement}\label{sec:accumulated-improvement}

\begin{figure}
    \centering
    \includegraphics[width=0.95\linewidth]{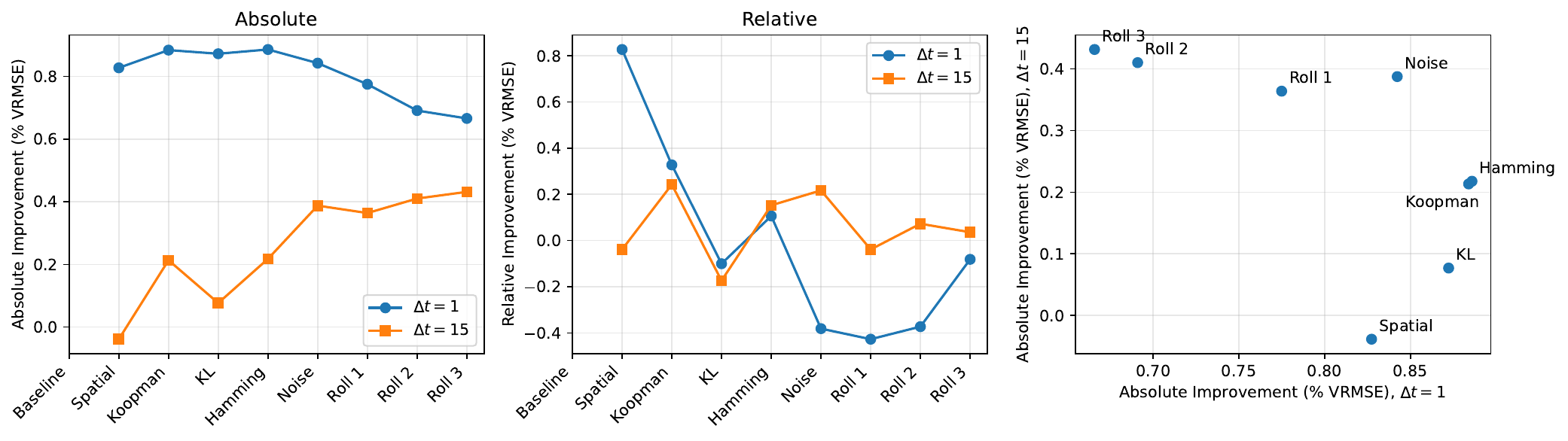}
    \caption{Summary of the cumulative improvement caused by each addition. Methods are included cumulatively on the x-axis (e.g., the `Hamming' model is trained using `Spatial', `Koopman', `KL', and `Hamming' methods). (a) Plot of the absolute percent improvement versus the baseline (the baseline would be $0$). (b) The relative improvement between each addition. (c) The correlation between the 1-step and 15-step absolute improvements.}
    \label{fig:accumulated_improvement}
\end{figure}

{
Although no single intervention dominates, these individual changes accumulate to produce a $43.2\%$ reduction in $15$-step rollout error relative to the vanilla LDM.
Table~\ref{tab:vrmse_by_dt} reports this accumulated improvement in VRMSE broken down by each addition, and these trends are summarized graphically in Figure~\ref{fig:accumulated_improvement}.
Most interestingly, the $1$-step and $15$-step rollouts tell opposite stories. 
The $15$-step rollout shows near-monotonic improvement from the changes considered, while the $1$-step rollout improves dramatically after the initial architectural modernization (the \emph{Spatial} row) and then degrades nearly monotonically with each subsequent addition. 
This inversion is not incidental; it follows directly from the mechanism established throughout this section.
One-step loss rewards exact, pointwise prediction, which favors autoencoder and dynamics-model precision, whereas the interventions that most improve long-rollout stability, decoder sensitivity reduction from the Koopman constraint, and robustness from both noise injection mechanisms, work specifically by making the model \emph{less} precise and more tolerant of small deviations.
This inverted correlation indicates that one step prediction (equivalently: the loss metric) is only a weak indicator of long-rollout performance. 

These improvements lead to competitive performance against reported errors for long rollouts, even without any architecture optimization. For example, comparing against the Well's initial target VRMSE, our results are competitive with their $1$-step rollout target ($0.1034$), but significantly exceed their $15$-step rollout target ($2.11$) \cite{ohana2024well}. The performance is comparable to Walrus and other foundation models (they report an average VRMSE over the first $20$ steps of $0.1262$, out performing ours, but the VRMSE rapidly rises to $1.24$ for steps $20$-$60$, showing worse long stability than ours) \cite{mccabe2025walrus}. However, we emphasize that this comparison is not entirely fair because it is made without any architecture optimization or scaling. Although absolute performance optimization was outside the scope of this paper, if desired, these can likely be pursued to exceed the performance of these foundation models, Appendix~\ref{app:active_matter_kwae_size}. 

Finally, considering Figure~\ref{fig:activematter_overview}, the nearly $40\%$ improvement is significantly more impactful than it seems: the resulting model retains stable, physically consistent rollouts through approximately $15$ steps, and although exact correspondence with the ground truth breaks down beyond this point, the model's predictions remain on the solution manifold rather than diverging into unrealizable states.
This distinction matters for downstream use: physically plausible predictions remain usable for further analysis even without pointwise accuracy, while physically implausible predictions generally are not, regardless of their VRMSE.
}

\subsection{Extrapolation beyond the training horizon: Application to high-cycle fatigue prediction}\label{sec:res-fatigue}

{
Our third and final question is whether the proposed LDM can extrapolate reliably far beyond the
horizon it was trained on. We test this question on the high-cycle fatigue case study, where failure
only emerges after $10^4$–$10^6$ loading cycles and simulating that many cycles directly is
prohibitively expensive (Section~\ref{sec:fatigue-casestudy}). If an LDM trained on a short window of
cycles can accurately extrapolate to the full fatigue life, it turns a simulation that takes weeks
into one that takes minutes.
}

{
We apply the LDM to the crystal-plasticity cyclic fatigue dataset introduced in
Section~\ref{sec:fatigue-casestudy}, reusing the training-level interventions established in
the previous Section with only minor additional tuning.
We performed limited optimization of the AViT architecture and found minimal improvement, Appendix~\ref{app:fatigue_avit_ablation_trends}.
We restricted training to
simulations of $2^9$ cycles or fewer, a small fraction of the long rollout horizon we ultimately
test against.
We trained two variants:
one that rolls out all $55$ physical fields jointly (Full), and
one specialized to predict only the Fatigue Indicator Parameter (FIP) field\footnote{They must be predicted directly, rather than estimated from the other fields because of the logarithmic time sampling.} \cite{rovinelli2015}.
We evaluated both variants on a held-out test set within the training horizon (up to $2^9$ cycles) and, more critically, on two long-time simulations extrapolated to $2^{15}$ cycles.
}

\begin{figure}
    \centering
    \includegraphics[width=\textwidth, height=0.67\textheight, keepaspectratio]{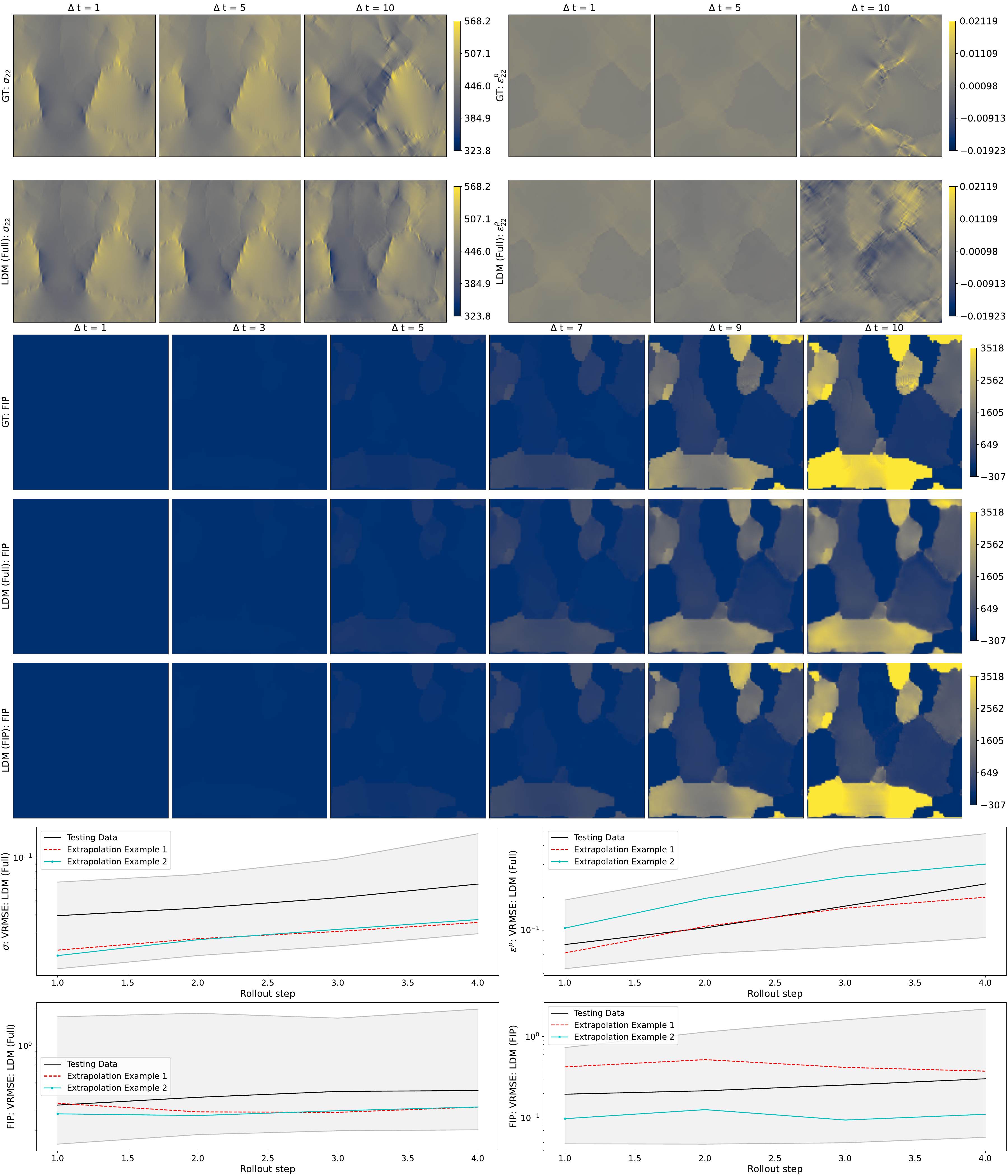}
    \caption{\textbf{Overview of proposed LDM performance on the Cyclic Fatigue Benchmark.} Rows 1,2 depict an example stress and strain field prediction. Rows 3-5 contrast predictions on an example Fatigue Indicator Parameter (FIP) field. Row 4 depicts the prediction from the LDM trained to predict all fields, Row 5 is the prediction from an LDM specialized to FIP predictions. All examples are derived from one of the expensive long horizon calculations performed. As a result, they depict time steps both within the training range and outside of it. Rows 6 and 7 report rollout errors over the training time distribution. Rollout errors for the two long-time examples are superimposed for context.}
    \label{fig:fatigue_overview}
\end{figure}

{
Figure~\ref{fig:fatigue_overview} summarizes performance across this full range.
In these predictions, the model is provided $5$ initial simulation snapshots as context, equivalent to $2^5$ cycles.
Subsequently, predictions are autoregressively generated.
Note that, in this setup, $\Delta t=4$ represents the boundary of training.
As observed in the first two rows of Figure~\ref{fig:fatigue_overview} the model continues making strong predictions of the stress and strain fields, even in extrapolation, with performance only severely degrading after significant extrapolation.
The VRMSE rollout errors (rows 6 and 7) for these fields indicates stable rollout and good predictions even up to the dataset's edge.
Looking at the results for the variant on FIP predictions (rows 3--5), we observe that the FIP specialized model outperforms the composite model.
The central result here is that, given only $2^5$ cycles of context, far inside the training regime, the model predicts the FIP at $2^{15}$ cycles, six orders of magnitude beyond the cycles it has seen, with less than $10\%$ error.
This level of extrapolation accuracy is the direct payoff of the training-level interventions because those interventions systematically lower decoder sensitivity and stabilize the latent dynamics, the model's error grows slowly enough with rollout length that a short training window still constrains behavior many cycles later. 
}

\begin{figure}
    \centering
    \includegraphics[width=0.7\linewidth]{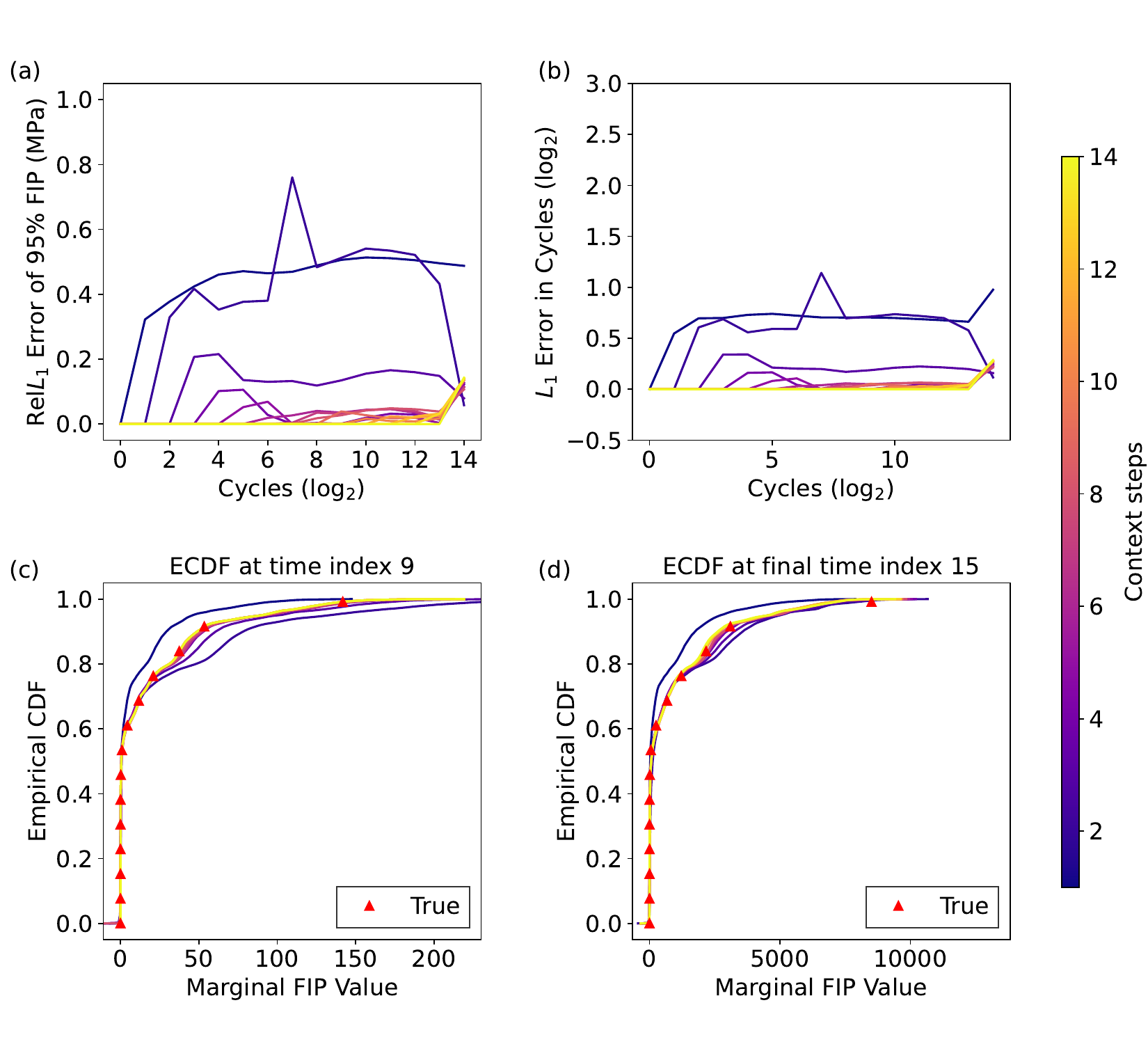}
    \caption{\textbf{Summary of cyclic fatigue based failure prediction on long rollout Example 1 using the LDM (Fatigue Indicator Parameter (FIP)) model.} In all plots, lines are colored by the number of simulation context steps provided before switching to the neural solver. (a) the relative $L_1$ error of the predicted $95$-percentile FIP at each cycle order of magnitude. (b) A reinterpretation of (a): $L_1$ error in the number of cycles to failure if failure occurred at each cycle order of magnitude. (c) predicted empirical FIP CDF at the temporal boundary of training data. (d) predicted Empirical FIP CDF at maximum ground truth cycle count ($2^{15}$ cycles).}
    \label{fig:fatigue_fip_predictions}
\end{figure}

Figure~\ref{fig:fatigue_fip_predictions} reports predictions for the first long-rollout example in
detail (see Appendix~\ref{appdx:fatigue-expt} for the second example).
We note that the quality of the prediction improves as more simulation context is provided (Figure~\ref{fig:fatigue_fip_predictions}a,c,d).
But even with limited context, the model remains accurate far beyond its training horizon.
Given just $2^5$ cycles of context, an easily achievable cost, the model predicts the $95$th-percentile FIP at $2^{15}$ cycles with less than $10\%$ error.
This accuracy holds across the full
distribution, not only the tail (Figure~\ref{fig:fatigue_fip_predictions}c,d), and we observe no
meaningful drop in predictive power between the edge of the training distribution ($2^9$ cycles) and
$2^{15}$ cycles.

Because these extreme-value statistics of the FIP field are commonly used to estimate the onset of failure, even small prediction error can matter.
Figure~\ref{fig:fatigue_fip_predictions}b puts this in context by reinterpreting field error as \emph{cycle} error: for each point on the x-axis,
the plot shows how far off the model's predicted failure cycle would be if failure occurred at that
point.
With only $2^5$ cycles of context from the crystal plasticity solver, the model localizes a potential
failure at $2^{15}$ cycles to within half an order of magnitude.
A result  accurate enough to be useful for early screening, long before a full simulation would be available.

\section{Discussion}\label{sec:discuss}

\subsection{Foundation Models}

\begin{figure}
    \centering
    \includegraphics[width=1.0\linewidth]{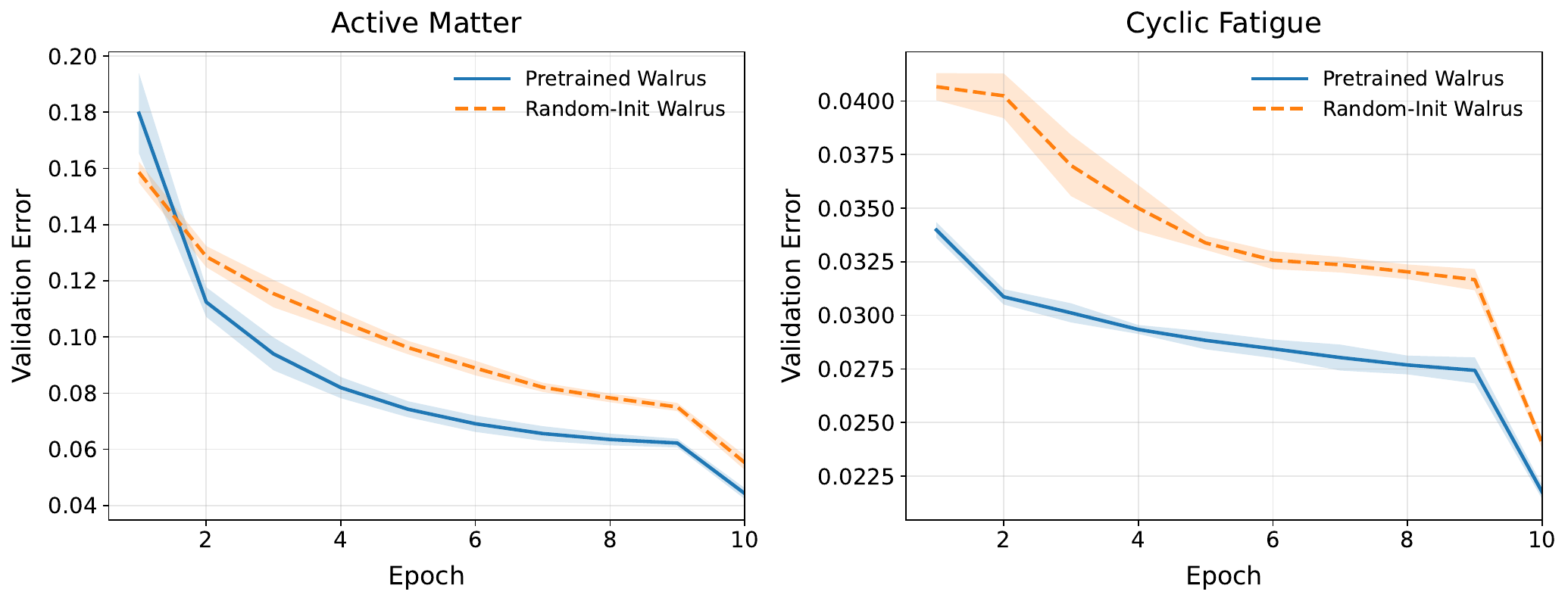}
    \caption{\textbf{Effect of foundation-model (Walrus) pretraining on validation error across benchmarks.}
    Validation error versus training epoch for Walrus initialized from its pretrained weights compared to a randomly initialized model of identical architecture, on the active matter (Sec.~\ref{sec:activematter-casestudy}) and high-cycle fatigue (Sec.~\ref{sec:fatigue-casestudy}) case studies. Shaded regions denote $\pm 1$ standard deviation across $10$ random seeds. The pretrained model reaches lower validation error consistently across both datasets and throughout training, despite two compounding distribution shifts: Walrus was pretrained on real-space physical fields, not the compressed latent encodings used here, and, in the fatigue case, on a dataset it never saw during pretraining at all.
    }
    \label{fig:pretrain_vs_random_walrus}
\end{figure}

As described in Section~\ref{sec:walrus-foundation-model}, we envision a beneficial alignment between these interventions and recent developments in physical foundation models because of the spatially structured latent encodings we have adopted. 
Here, we analyze the impact on training of initializing the dynamic model using a pre-trained foundation model. 
Specifically, we compare fine-tuning a pretrained Walrus checkpoint against a randomly initialized Walrus of identical architecture under a matched computational budget (fixed epochs, compute nodes, and GPUs).
We analyze both the active matter and high-cycle fatigue case studies.
Figure~\ref{fig:pretrain_vs_random_walrus} shows the resulting validation curves.
Notably, Walrus was pretrained on real-space data from the Well's active matter dataset, but had never encountered the high-cycle fatigue data in any form.
In both cases, the pretrained model reaches a low validation error within a small fraction of the epochs the randomly initialized model needs to reach the same error, and it maintains this advantage for the remainder of fine-tuning.
This faster convergence holds even for high-cycle fatigue, a system Walrus never saw during pretraining, indicating that the advantage comes from transferable structural priors rather than memorized features of the active matter dataset.
While limited in scope, this experiment points to a promising advantage of image-like latent spaces: they preserve enough spatial structure for an LDM to draw on the transferable representations already learned by existing physical foundation models, substantially reducing the training cost of adopting one. 

\section{Conclusion}\label{sec:conclusion}

This paper presents a systematic study of training-level interventions for LDMs, showing that rollout instability is not an inherent cost of operating in a compressed latent space, but a consequence of how that space is conventionally trained. 
Two mechanisms drive the improvements we observe.
First, lighter compression alone raises the floor on single-step accuracy to a level comparable with pixel-space models.
Second, and more importantly, reducing \emph{decoder sensitivity}, the degree to which small latent-space errors are amplified into large errors in the reconstructed field, is the primary lever for long-rollout stability, since it is what determines how much a rollout's accumulated latent-space error is projected into the solution space.
The Koopman-inspired dynamics constraint and Hamming noise injection both act specifically on this mechanism, each independently reducing decoder sensitivity and improving long-rollout accuracy.
Importantly, these gains are not visible from single-step accuracy alone, see Section~\ref{sec:accumulated-improvement}.
Interventions that improve long-rollout stability often degrade direct loss metrics (such as single step prediction error).
This major conclusion from this study highlights that task misalignment exists in current models and emphasizes that long-rollout stability must be optimized for directly rather than assumed to follow from standard reconstruction or dynamics losses.
In three case studies, we see several benefits applying the proposed LDM.
The spinodal decomposition benchmark demonstrates the LDM's beneficial cost-accuracy trade-off: the LDM matches the performance of established pixel space models at two orders of magnitude lower computational cost and half the required VRAM.
In the more dynamic active matter study, the proposed LDM improves long rollout stability.
Applied to a high-cycle fatigue case study, the same framework predicts failure with less than $10\%$ error at cycle counts six orders of magnitude beyond the training horizon, demonstrating that these training-level interventions generalize from controlled benchmarks to a demanding extrapolation task.
By treating latent space as a dynamical design variable rather than a passive compression bottleneck, neural surrogates can be trained to remain stable, efficient, and useful over the long horizons that scientific prediction demands.

\section*{Acknowledgments}
A.E.R., B.A.J., K.G., and R.D. are supported by the U.S\@.~Department of Energy, Office of Science, Office of Advanced Scientific Computing Research and  Office of Basic Energy Sciences, Scientific Discovery through Advanced Computing (SciDAC) program under the MIRAGE project.
V.O. and J.D.S. are supported by the Laboratory Directed R\&D program at Sandia National Laboratories.
A.T.L. and D.L.D. are supported by the DoD/DOE Joint Munitions Program.
Computational capabilities used to performed the work were developed by the Center for Integrated Nanotechnologies (CINT), an Office of Science user facility operated for the U.S\@.~Department of Energy.
This article has been authored by an employee of National Technology \& Engineering Solutions of Sandia, LLC under Contract No\@.~DE-NA0003525 with the U.S\@.~Department of Energy (DOE).
The employee owns all right, title, and interest in and to the article and is solely responsible for its contents. The United States Government retains and the publisher, by accepting the article for publication, acknowledges that the United States Government retains a non-exclusive, paid-up, irrevocable, world-wide license to publish or reproduce the published form of this article or allow others to do so, for United States Government purposes.
The DOE will provide public access to these results of federally sponsored research in accordance with the DOE Public Access Plan {https://www.energy.gov/downloads/doepublic-access-plan.}

\appendix
\numberwithin{equation}{section}

\section{Additional dataset details}\label{apdx:dataset}
\subsection{Spinodal decomposition}
We used the spinodal decomposition phase-field dataset used in Dingreville et al.~\cite{dingreville2024benchmarking} and generated following the PFHub Cahn--Hilliard benchmark problem~\cite{wheeler2019}.
The total free energy is given as follows.
\begin{equation}
    F[c] = \int_{\Omega}
    \left[
    f_{\mathrm{bulk}}(c)
    +\frac{\kappa}{2}|\nabla c|^2
    \right]dV.
\end{equation}
It uses a symmetric double-well bulk energy.
\begin{equation}
    f_{\mathrm{bulk}}(c) = 
    W(c-c_{\alpha})^2(c-c_{\beta})^2,
    \quad
    c_{\alpha}=0.3,\quad c_{\beta}=0.7.
\end{equation}
Combining these, the conserved concentration field evolves according to the following expression.
\begin{equation}
    \frac{\partial c}{\partial t}=
    \nabla\cdot
    \left[
    M\nabla
    \left(
    \frac{\partial f_{\mathrm{bulk}}}{\partial c}
    -\kappa\nabla^2c
    \right)
    \right].
\end{equation}
It is subject to periodic boundary conditions and the following initial condition.
\begin{equation}
    c(\mathbf{x},0)=c^*+A\zeta(\mathbf{x}).
\end{equation}
\noindent Here, $\zeta$ is a random perturbation. The equations were solved using a first-order semi-implicit Fourier-spectral scheme.
The dataset contains 1\,000 simulations generated by Latin hypercube sampling of four conditioning variables: the average concentration $c^*\in[0.39,0.61]$, gradient-energy coefficient $\kappa\in[0.75,3.5]$, mobility $M\in[3.7,6.5]$, and barrier height $W\in[3,8.5]$.
Each simulation tracks the single scalar concentration field $c(x,y,t)$ on a $384\times384$ grid, evolved from $t=0$ to $t=10\,000$ with 201 recorded states including the initial condition; the released learning arrays contain 200 spatial frames per trajectory.
Following the procedure from the original dataset, 700 simulations were used for training, 200 for validation, and 100 for testing.

\subsection{Active matter}
We used the active matter dataset from the Well benchmark datasets, which models a dense suspension of active particles in an incompressible Stokes fluid~\cite{maddu2024learning, ohana2024well}.
The particle distribution $\Psi(\mathbf{x},\mathbf{p},t)$ evolves according to the Smoluchowski equation
\begin{equation}
    \frac{\partial\Psi}{\partial t}
    +\nabla_{\mathbf{x}}\cdot(\dot{\mathbf{x}}\Psi)
    +\nabla_{\mathbf{p}}\cdot(\dot{\mathbf{p}}\Psi)=0,
\end{equation}
with conformational fluxes
\begin{align}
    \dot{\mathbf{x}}
    &=\mathbf{u}-d_T\nabla_{\mathbf{x}}\log\Psi,\\
    \dot{\mathbf{p}}
    &=(\mathbf{I}-\mathbf{p}\mathbf{p})
      \cdot(\nabla\mathbf{u}+2\zeta\mathbf{D})
      \cdot\mathbf{p}
      -d_R\nabla_{\mathbf{p}}\log\Psi .
\end{align}
Here, $\mathbf{D}=\langle\mathbf{p}\mathbf{p}\rangle$ is the second orientational moment and $\zeta$ controls steric alignment.
The particle dynamics are coupled to incompressible Stokes flow,
\begin{align}
    -\Delta\mathbf{u}+\nabla P
    &=\nabla\cdot\boldsymbol{\Sigma},
    &
    \nabla\cdot\mathbf{u}
    &=0,\\
    \boldsymbol{\Sigma}
    &=\alpha\mathbf{D}
      +\beta\,\mathbf{S}:\mathbf{E}
      -2\zeta\beta
      \left(\mathbf{D}\cdot\mathbf{D}-\mathbf{S}:\mathbf{D}\right),
\end{align}
where $\mathbf{E}=(\nabla\mathbf{u}+\nabla\mathbf{u}^{T})/2$ and $\mathbf{S}=\langle\mathbf{p}\mathbf{p}\mathbf{p}\mathbf{p}\rangle$.
Active stresses drive the fluid, while the resulting flow transports and reorients the particles, producing nonlinear and nonlocal feedback; because the evolution of the stored second moment depends on the unresolved fourth moment, the system also contains the implicit closure problem discussed in Section~\ref{sec:datasets}.
The dataset stores time series of the concentration $c$, velocity $\mathbf{u}$, orientation tensor $\mathbf{D}$, and strain-rate tensor $\mathbf{E}$ as primary target variables, along with the conditioning variables $\alpha$, $\beta$, and $\gamma$ for each series.
The complete dataset contains 225 time series, each spanning 81 time steps.

\subsection{Crystal plasticity high-cyclic fatigue}
{
The third dataset models high-cycle fatigue damage accumulation in polycrystalline aluminum, generated from microstructures extracted from a 3D experimental sample measured via diffraction contrast tomography (Appendix~\ref{apdx:polymicrosextension}, Figure~\ref{fig:polymicros_periodic_boundaryconditions}).
}

\begin{figure}
    \centering
    \includegraphics[width=0.5\linewidth]{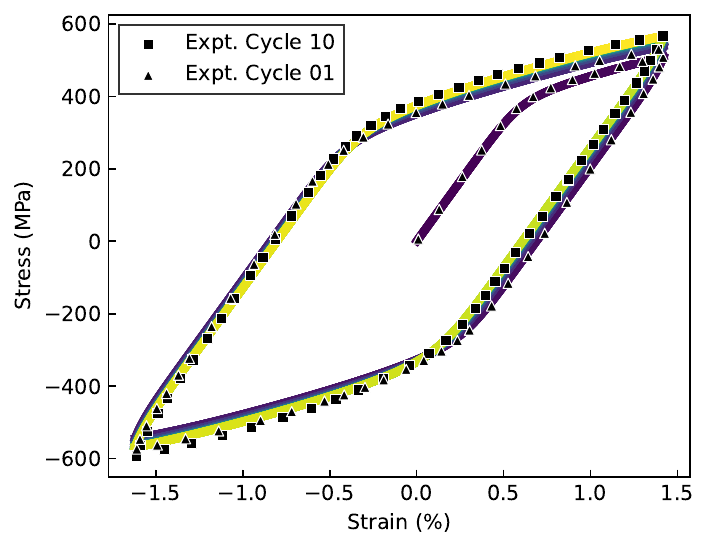}
    \caption{\textbf{Calibration of the crystal plasticity model against experimental cyclic loading data.} Simulated stress--strain response using the parameters in Table~\ref{tab:cp-params}, compared against experimental data for aluminum~\cite{wang2021}. Simulated curves are shaded from dark to light with increasing cycle number, tracking the model's evolving hardening response across cycles.}
    \label{fig:cp-params}
\end{figure}

{
Cyclic fatigue simulations were performed using MASSIF, a parallel crystal-plasticity fast Fourier transform (CPFFT) code~\cite{tari2018,lebensohn2020spectral}, following the modeling framework of Asaro~\cite{asaro1983}, via a spectral method~\cite{lebensohn2020spectral}.
A multiplicative decomposition of the deformation gradient represents the plastic velocity gradient in terms of slip-rate components on each slip system, governed by a rate-dependent flow rule,
\begin{equation}
    \dot{\gamma}^\alpha = \dot{\gamma}_0 \left\langle \frac{\tau^\alpha - \chi^\alpha}{\tau_c^\alpha} \right\rangle ^{1/m} \mathrm{sgn} \left( \tau^\alpha - \chi^\alpha\right),
    \label{eqn:powerlaw}
\end{equation}
where $\dot{\gamma}^\alpha$ is the slip rate on system $\alpha$, $\dot{\gamma}_0$ is a reference slip rate, $m$ is the strain rate sensitivity, $\tau^\alpha$ is the resolved shear stress on system $\alpha$, $\tau_c^\alpha$ is the slip system strength, $\chi^\alpha$ is the backstress, and the Macaulay brackets represent the max of zero or their argument.
Isotropic hardening follows a Voce rule, $\tau^\alpha_c = \tau_0 + (\tau_1 + \theta_1 \Gamma)(1 - \exp[\Gamma|\theta_0/\tau_1|])$, where $\Gamma$ is the summed slip over all systems, and kinematic hardening follows a single term of the Ohno--Wang evolution rule~\cite{ohno1993}, a modification of the Armstrong--Frederick formulation~\cite{frederick2007}:
\begin{equation}
    \dot{\chi}^\alpha = h_1 \dot{\gamma}^\alpha - r_1 \chi^\alpha \left( \frac{\chi^\alpha}{h_1/r_1} \right)^{d_1} \left| \dot{\gamma}^\alpha \right|
    \label{eqn:backstress}
\end{equation}
with $h_1$, $r_1$, and $d_1$ as material parameters.
The fatigue indicator parameter (FIP) is the Fatemi--Socie parameter~\cite{fatemi1988critical}, or ``E5'' in the lexicon of Rovinelli \textit{et al.}~\cite{rovinelli2015},
\begin{equation}
    \text{FIP}_{\text{E5}} = \max_p \sum_{\alpha=1}^{N_s}\left| \tau_p^\alpha\Gamma_p^\alpha \right| \left( 1 + k \frac{\sigma_n^p}{\sigma_Y} \right)
    \label{eqn:fip}
\end{equation}
with $p$ indicating slip planes according to which the plane normal stress $\sigma_n^p$ is considered in relation to $k$ and $\sigma_Y$, two additional material parameters.
Material parameters were calibrated against a fully reversed uniaxial tension dataset for aluminum 7075-T6~\cite{wang2021}; Figure~\ref{fig:cp-params} and Table~\ref{tab:cp-params} show the resulting fit and calibrated values, which show good agreement with the experimental data for both the initial and tenth loading cycle.
}

{
The dataset contains $491$ simulations. $341$ simulations were used for training and $75$ were used for each of validation (used in hyperparameter tuning) and testing (used in the paper). All three datasets contained simulations of varying lengths of time. For example, the training dataset contained $8$ simulations at $2^6$ cycles, $32$ at $2^7$, $151$ at $2^8$, and $150$ at $2^9$. 
Each training simulation draws a modification factor from a normal distribution ($\mu=1$, $\sigma=0.2$) for each material parameter, a $128 \times 128$ microstructural slice drawn without replacement from the periodically padded dataset, and a cyclic loading condition sampled from target distributions for mean strain (lognormal, $\mu=0\%$, $\sigma=1\%$) and strain amplitude (uniform, $0.5\%$ to $5\%$).
At the peak of each loading cycle, for log-spaced cycle numbers, we record the full spatial fields for FIP, backstress, total and plastic strain, Cauchy stress, Euler angles, and critical resolved shear strength, resulting in $55$ co-evolving field quantities per snapshot.

}

\begin{table}
    \caption{\textbf{Crystal-plasticity parameters.} Calibrated parameter values for fully reversed loading using an example microstructure from the present work and mechanical data from Ref. \cite{wang2021}.}
    \centering
    \begin{tabular}{l r r}
        \hline
        Parameter & Value & Unit\\
        \hline
        $\tau_{0}$   & 175   & MPa\\
        $\tau_{1}$   & 25    & MPa\\
        $\theta_{0}$ & 200   & MPa\\
        $h_1$     & 7000  & MPa\\
        $r_1$     & 100   & MPa\\
        $d_1$     & 4     & --\\
        $\dot{\gamma}_0$ & 1.0 & --\\
        $m$ & 8 & --\\
        $k$ & 1.0 & --\\
        $\sigma_Y$ & 200 & MPa\\
        \hline
    \end{tabular}
    \label{tab:cp-params}
\end{table}

\subsubsection{Periodic microstructure extension using PolyMicros}
\label{apdx:polymicrosextension}
{
The experimentally collected polycrystalline microstructures used in Case Study 3 were periodically extended to be compatible with the MASSIF CPFFT solver using the PolyMicros foundation model \cite{buzzy2026polymicros} via a custom periodic extension module built on
PolyMicros' post-training conditioning capability.
}

{
We extracted 3D cubes from the experimental microstructure, with each edge spanning approximately $10$ grains, and periodically extended each cube using the diffusion-based conditioning process of Robertson \textit{et al.}~\cite{robertson2023local}:
the original $64$-pixel patch is embedded in an $82$-pixel periodic domain, and the ground-truth patch is reintroduced at every diffusion step for the first $80\%$ of the process, then left uninterrupted for the remaining $20\%$, balancing fidelity to the original data with a smooth periodic transition.
The resulting volume was resampled to $128\times128\times128$ via nearest-neighbor interpolation and sliced along all three axes at $10$-pixel intervals, larger than a grain, to limit correlation between extracted 2D slices.
}

\begin{figure}
    \centering
    \includegraphics[width=0.7\linewidth]{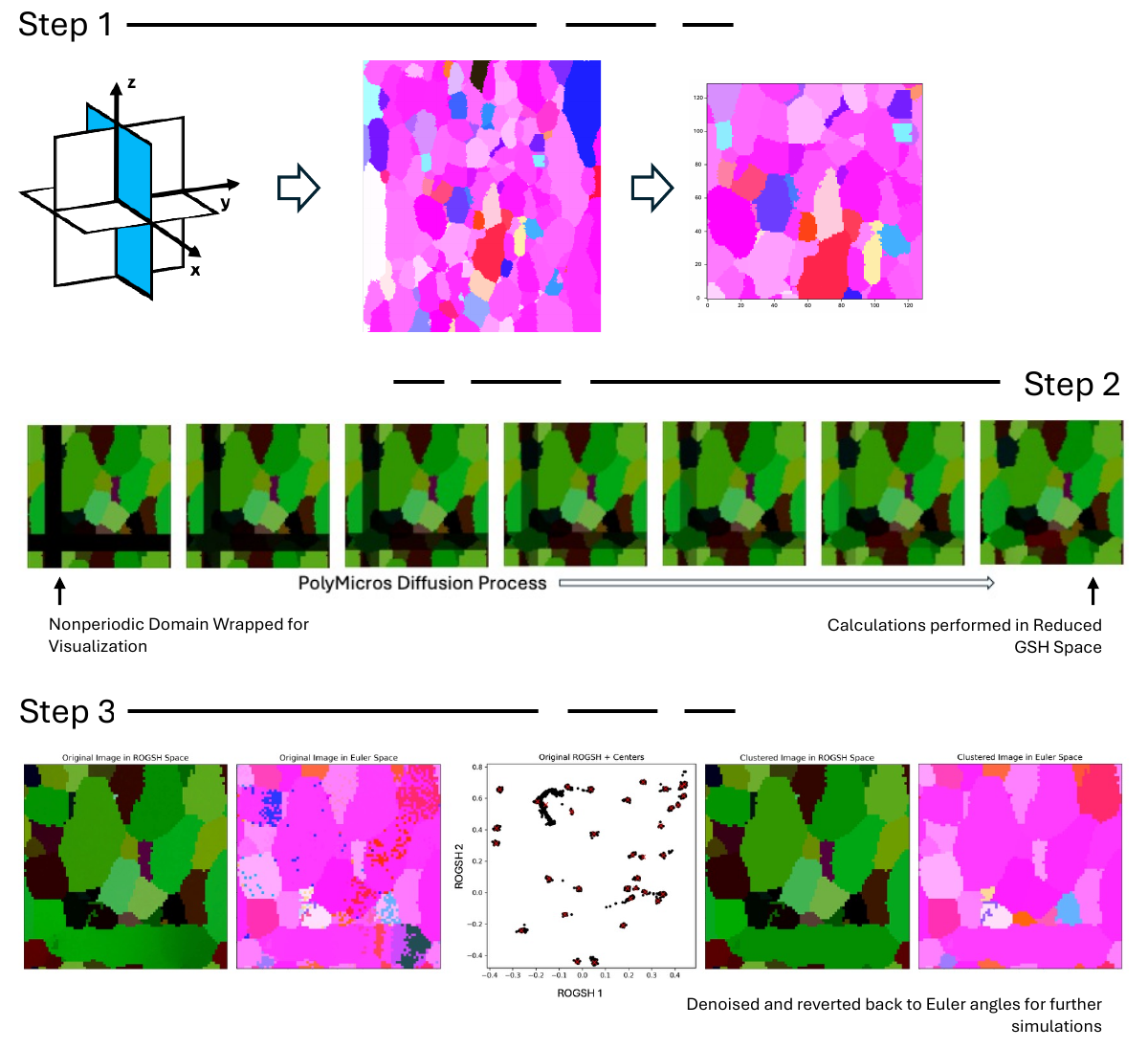}
    \caption{\textbf{Overview of the microstructure preparation pipeline.}
    Step 1: individual 2D slices are extracted from the 3D DCT-measured aluminum microstructure along each principal direction.
    Step 2: each slice is periodically extended into a $128 \times 128$ patch to accommodate the CPFFT simulation code, using a periodic-extension conditioning process, using the PolyMicro Foundation Model~\cite{buzzy2026polymicros}.
    Step 3: Step 3: pixel values are transformed from PolyMicro's ROGSH basis back to Euler angle space.}
\label{fig:polymicros_periodic_boundaryconditions}
\end{figure}

After padding, pixel values were then transformed from PolyMicros' Reduced Order Generalized Spherical Harmonics (ROGSH) basis back to Euler angle space using a k-dimensional-tree-based nearest-projection strategy against a reference set of Euler angle--ROGSH pairs, following Buzzy \textit{et al.}~\cite{buzzy2024statistically}.
This projection introduces visual artifacts (Figure~\ref{fig:polymicros_periodic_boundaryconditions}, Row 3), because boundaries in the Euler angle fundamental zone are irregular in a way the ROGSH space is not: points far apart under the Euler angle metric can map close together in ROGSH space.
These artifacts are cosmetic rather than physical, since the affected pixels remain crystallographically equivalent under the symmetries of Euler space, but we remove them by clustering the ROGSH space with Density-Based Spatial Clustering of Applications with Noise (DBSCAN) and replacing each cluster's pixels with its mean value, a purely spatial-structure-agnostic cleanup that largely eliminates the artifacts.

\section{Model architectures}\label{apdx:architecture}
{
This appendix provides the full architectural and training specifications for the three models introduced in Section~\ref{sec:baseline-ldm}:
the UNet baseline,
the axial-vision transformer (AViT) dynamics model, and
the wavelet autoencoder.
All architectures were trained for 200\,000 gradient steps using a learning rate of $3 \times 10^{-4}$, the SOAP optimizer~\cite{vyas2025soap} with a weight decay of $1 \times 10^{-5}$, and all other optimizer parameters left at their default values.
We used a two-stage learning rate schedule: 5\,000 steps of linear warmup followed by cosine annealing over the remaining 195\,000 steps.
These values were not optimized, but were observed to perform well across all three architectures.
}

\subsection{UNet}
{
We adopted the UNet architecture described by Dingreville et al.~\cite{dingreville2024benchmarking}, a variant of the convolutional network originally introduced for denoising diffusion probabilistic models (DDPMs)~\cite{ho2020denoising}, repurposed here simply as a general-purpose image-to-image network rather than for diffusion.
Following Dingreville et al. and Oommen et al.~\cite{dingreville2024benchmarking, oommen2022learning}, we trained the network using a bundled strategy:
rather than predicting one time step at a time, the network is given the 5 preceding time steps as input and is trained to directly predict the following 5 steps as output.
Training uses a direct $L_1$ loss between the 5 predicted steps and the ground truth.
}
\subsection{Axial Vision Transformer (Axial-ViT)}
{
We used an axial-vision transformer (AViT) inspired by the multiple physics pretraining (MPP) architecture~\cite{mccabe2024multiple, ho2019axial}, which factorizes full spatiotemporal attention into two alternating passes: a temporal attention mechanism, operating on every time step of each spatial pixel individually, and a spatial attention mechanism, operating on the entire spatial field for a single time step.
Unlike MPP, we used dense attention for the spatial mechanism, meaning every spatial position attends to every other position rather than attending over a single dimension, and a patch size of just 1, meaning each latent cell is treated as its own token rather than being grouped with neighboring cells; both are made tractable by operating on the autoencoder's compressed latent representation rather than the full-resolution field.
We used the pre-norm transformer layout~\cite{xiong2020layer}, in which normalization is applied before, rather than after, each attention and MLP sublayer for improved training stability, with dropout of 0.1 applied in both the attention mechanism and the subsequent MLP, and the MLP itself is a standard two-layer network using the ReLU$^2$ activation function~\cite{so2021searching}.
We imposed causal masking on the temporal attention mechanism, preventing the model from attending to future time steps when making a prediction, and use rotary position embeddings (RoPE) to encode positional order in both the temporal and spatial attention mechanisms~\cite{su2024roformer}; for the spatial mechanism, we use Qwen's multimodal RoPE (MRoPE), an extension of RoPE that encodes two positional axes independently, to account for both $x$ and $y$ spatial dependence~\cite{huang2025revisiting}.
The AViT is pretrained using standard teacher forcing, in which the network is given ground-truth context rather than its own predictions at each step: for the Active Matter problem, the network is provided $9$ time steps of context, and the updated output is penalized with an $L_1$ loss against 9 shifted target time steps, with earlier steps given less context due to the imposed causal mask for temporal attention.
For dynamics-side noise injection (Section~\ref{sec:interventions}), white noise is added to the AViT's input before making predictions.
During rollout fine-tuning, the pretrained network is further trained by feeding its output block back into the network a fixed number of times, with the loss computed only on the final output; this forces the network to learn to account for its own mistakes (Appendices~\ref{apdx:active_matter_avit_ablation} and~\ref{app:fatigue_avit_ablation_trends}).

We trained the dynamics models using a batch size of 16 on a single A6000 GPU, with gradient accumulation used during rollout fine-tuning.
}

\subsection{Wavelet autoencoder}
{
We used an inhouse implementation of the LiteVAE wavelet autoencoder~\cite{sadat2024litevae}. The LiteVAE architecture combines a wavelet encoder with DDPM blocks~\cite{ho2020denoising} and a stable-diffusion-type decoder~\cite{rombach2022high}, adopted for its demonstrated training efficiency and its tendency to promote interpretable spatial structuring in the latent space~\cite{sadat2024litevae}.
The full autoencoder, including both Koopman learning and the KL penalty, is trained with the following composite loss,
\begin{equation}
    \mathcal{L}_\theta = 20.0 \, \|u_t - \hat{u}_t\|_1 + 10.0 \, \|u_{t+1} - \hat{u}_{t+1}\|_1 + 0.001 \, \|K\| + 0.1 \, \mathcal{L}_{KL},
    \label{eq:composite-loss}
\end{equation}
where
\begin{equation}
    \hat{u}_t = g_\psi\big(f_\phi(u_t) + 0.01\,\epsilon\big), \qquad
    \hat{u}_{t+1} = g_\psi\big(K(f_\phi(u_t) + 0.01\,\epsilon)\big),
    \label{eq:composite-loss-terms}
\end{equation}
$g_\psi(\cdot)$ and $f_\phi(\cdot)$ are the model's decoder and encoder as defined in Section~\ref{sec:baseline-ldm}, $\epsilon$ is white noise, and $K$ is the Koopman operator introduced in Section~\ref{sec:interventions}, implemented as a bias-free, kernel-$3$ convolution.
The term $\|K\|$ penalizes the magnitude of the convolution kernel, following the recommendation of Geneva and Zabaras to prevent arbitrary rescaling of the latent space~\cite{geneva2022transformers}.
Finally, $\mathcal{L}_{KL} = \left\langle (\left\langle z_t \right\rangle_b)^2 \right\rangle_c + \left\langle \left( \left\langle (z_t - \left\langle z_t \right\rangle_b)^2 \right\rangle_b - 1.0\right)^2 \right\rangle_c$, matching the definition in Section~\ref{sec:interventions}, where $\left\langle \cdot \right\rangle_b$ denotes the mean over the batch and $\left\langle \cdot \right\rangle_c$ the mean over all remaining components.
This term encourages the latent space to adopt a unit Gaussian structure.
All loss calculations are performed on mean-standard-deviation normalized variables.
We trained the autoencoder using a batch size of $32$ on a single A6000 GPU.
}

\section{Estimating decoder sensitivity}\label{apdx:sensitivity}
{
A autoregressive dynamics model inevitably makes mistakes when autoregressive rollout is performed.
In a LDM, these perturbations are passed to the decoder.
Therefore, the sensitivity of the decoder with respect to the latent encoding directly defines how much error is propagated back into the problem's solution space.
As shown in the main experiments in the core of the paper, insensitive decoders offer the opportunity for controlling error propagation. 
}

{
Interpreting the sensitivity of the autoencoder latent space requires understanding how strongly the decoder uses each latent coordinate.
Given a decoder $f_\theta : \mathbb{R}^{d_z} \to \mathbb{R}^{d_o}$ with latent encoding \(z \in \mathbb{R}^{d_z}\), we define the coordinate-wise root-mean-square sensitivity of the decoder to latent dimension \(j\) as
\begin{equation}
    s_j(z)
    =
    \left(
    \frac{1}{d_o}
    \sum_{i=1}^{d_o}
    \left(
    \frac{\partial f_{\theta,i}(z)}{\partial z_j}
    \right)^2
    \right)^{1/2}.
\end{equation}
Equivalently, if $J(z) = \frac{\partial f_\theta(z)}{\partial z} \in \mathbb{R}^{d_o \times d_z}$, then the expression can be rewritten as:
\begin{equation}
    s_j(z)
    =
    \left(
    \frac{1}{d_o}
    \left[J(z)^\top J(z)\right]_{jj}
    \right)^{1/2}.
\end{equation}
}

\noindent Large values of \(s_j(z)\) indicate that small perturbations in latent coordinate \(j\) produce large local changes in the decoded output, while small values suggest that the decoder is locally insensitive to that coordinate. Averaging \(s_j(z)\) over a dataset gives a useful empirical measure of the extent to which latent dimensions are actively used by the decoder.

For high-dimensional outputs, explicitly forming the full Jacobian \(J(z)\) can be computationally expensive. Instead, we use a Hutchinson-style randomized estimator \cite{hutchinson1989stochastic,bekas2007estimator}. Let \(v \in \mathbb{R}^{d_o}\) be a random probe vector with independent Rademacher entries, \(v_i \in \{-1,+1\}\), so that
\begin{equation}
    \mathbb{E}[v] = 0,
    \qquad
    \mathbb{E}[vv^\top] = I.
\end{equation}

\noindent The gradient of the scalar projection \(v^\top f_\theta(z)\) is
\begin{equation}
    \nabla_z \left(v^\top f_\theta(z)\right)
    =
    J(z)^\top v.
\end{equation}

\noindent Therefore,
\begin{equation}
    \mathbb{E}_v
    \left[
    \left(J(z)^\top v\right)
    \odot
    \left(J(z)^\top v\right)
    \right]
    =
    \operatorname{diag}\left(J(z)^\top J(z)\right).
\end{equation}

\noindent Here, \(\odot\) denotes elementwise multiplication. Using \(K\) independent random probes \(v^{(k)}\), we can efficiently estimate the desired expression using a Monte Carlo estimator.
\begin{equation}
   \widehat{\operatorname{diag}\left(J^\top J\right)}
    =
    \frac{1}{K}
    \sum_{k=1}^{K}
    \left(J^\top v^{(k)}\right)
    \odot
    \left(J^\top v^{(k)}\right).
\end{equation}

Hence, we getting the following expression.
\begin{equation}
   \hat{s}_j(z)
    =
    \left(
    \frac{1}{d_o}
    \widehat{\left[J^\top J\right]_{jj}}
    \right)^{1/2}.
\end{equation}

This estimator avoids materializing the full Jacobian and requires only \(K\) vector-Jacobian products, which can be computed efficiently by reverse-mode automatic differentiation. Its cost is therefore controlled by the number of probes rather than by explicitly looping over all output components, making it practical for the large image-like data we encounter in this paper. We use 32 probes for the experiments discussed in this paper.

\section{Additional experiments}\label{apdx:add-exp}
\subsection{Spinodal decomposition}

Table~\ref{tab:relmse_by_dt_phasefield} reports the RelMSE for predictions for the spinodal decomposition benchmark, while Table~\ref{tab:vrmse_by_dt_phasefield} reports the VRMSE.
Calculations are performed at specific rollouts.
The rollout refers to the number of accumulated, autoregressive steps performed before calculating the loss.
Results reported in these Tables provide a baseline comparison to the active matter benchmark problem and the remaining Well benchmarks.

\begin{table}[ht]
\centering
\caption{\textbf{Average ReLMSE for predictions for the spinodal decomposition benchmark for various models and rollouts $\Delta t$.} }
\label{tab:relmse_by_dt_phasefield}
\begin{tabular}{lcccc}
\hline
Model & $\Delta t=1$ & $\Delta t=5$ & $\Delta t=15$ & $\Delta t=40$ \\
\hline
UNET & 0.00003 & 0.00024 & 0.00167 & 0.00598 \\
LDM & 0.00016 & 0.00051 & 0.00203 & 0.00586 \\
Vanilla LDM & 0.00261 & 0.02099 & 0.08853 & 0.18081 \\
\hline
\end{tabular}
\end{table}

\begin{table}[ht]
\centering
\caption{\textbf{Average VRMSE for predictions for the spinodal decomposition benchmark for various models and rollouts $\Delta t$.}}
\label{tab:vrmse_by_dt_phasefield}
\begin{tabular}{lcccc}
\hline
Model & $\Delta t=1$ & $\Delta t=5$ & $\Delta t=15$ & $\Delta t=40$ \\
\hline
UNET & 0.01274 & 0.02715 & 0.17213 & 0.13762 \\
LDM & 0.04619 & 0.04988 & 0.18177 & 0.14900 \\
Vanilla LDM & 0.90402 & 1.11307 & 1.27472 & 1.28404 \\
\hline
\end{tabular}
\end{table}

\subsection{Active matter}

In this appendix, we summarize additional information from experiments on the active matter dataset.

\subsubsection{Latent space size}
\label{app:active_matter_kwae_size}

\begin{figure}
    \centering
    \includegraphics[width=1.0\linewidth]{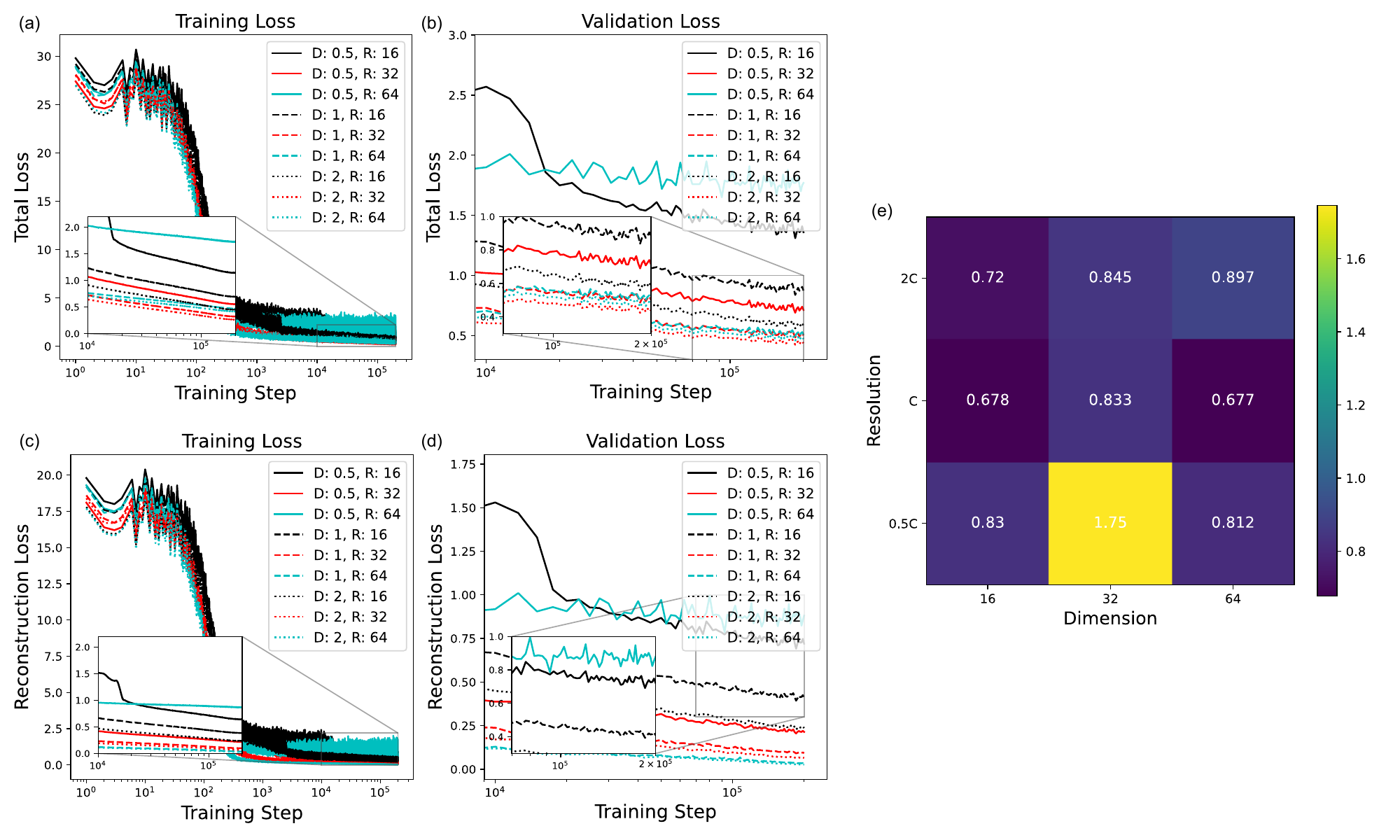}
    \caption{\textbf{Optimization of autoencoder hyperparameters for reconstruction weakens rollout performance.}
    (a-d) The reconstruction and total loss of the autoencoder swept over different latent space resolutions and channel numbers.
    (e) the corresponding $15$-step rollout error for the same hyperparameter sweep. Note the lack of correlation between trends in each.}
    \label{fig:latentspacesize_ablation}
\end{figure}

We perform edexperiments to understand the impact of the latent space size (both the number of latent channels and the resolution) on reconstruction accuracy and downstream multi-step rollout performance (here, we report 15 steps).
We observe stronger trends connecting latent space size to reconstruction accuracy.
Figure~\ref{fig:latentspacesize_ablation}a-d illustrates that increasing the number of channels consistently improves the reconstruction performance. 
However, the improvement saturates with minimal improvement being observed when moving from a number of latent channels equal to the number of ambient channels to a 2:1 ratio. 
In contrast, performance is optimized with an intermediate resolution, $32$. 

Interestingly, these clear trends do not translate cleanly into performance improvements in multi-step rollout, see Figure~\ref{fig:latentspacesize_ablation}e.
Here, we see that by optimizing for reconstruction, we actually selected suboptimal parameters for rollout. 
This, again, re-emphasizes a main conclusion from the main body: single step performance is only a weak indicator of neural surrogate solver performance. 

We also observe that the interventions involving noise injection were more effective when the latent space size was larger, see Table~\ref{tab:vrmse_channel_size}.
This indicates that these interventions can be further optimized by adjusting the architecture size if a specific performance is desired.

\begin{table}[ht]
\centering
\caption{
\textbf{Sensitivity of cumulative intervention gains to latent channel count.}
Average VRMSE at rollout $\Delta t=15$ on the active matter benchmark, comparing an $8$-channel and an $11$-channel latent space. Each row adds one training-level intervention cumulatively on top of all rows above it, in the order introduced in Section~\ref{sec:interventions}. \emph{Error} reports the raw VRMSE for that row's model; \emph{Absolute} reports its improvement over the baseline (row 1).}
\label{tab:vrmse_channel_size}
\begin{tabular}{|lcc|cc|}
\hline
Model & \multicolumn{2}{c|}{Channels: $8$} & \multicolumn{2}{c|}{Channels: $11$} \\
\cline{2-5}& Error & Absolute & Error & Absolute \\
\hline
Baseline & 0.98166 & -- & 0.98166 & -- \\
Hamming noise & 0.76793 & 0.21773 & 0.73952 & 0.24667 \\
Dynamic noise & 0.60120 & 0.38757 & 0.57556 & 0.41368 \\
Roll 1 & 0.62439 & 0.36394 & 0.55968 & 0.42987 \\
Roll 2 & 0.57895 & 0.41024 & 0.51364 & 0.47677 \\
Roll 3 & 0.55802 & 0.43156 & 0.47249 & 0.51868 \\
\hline
\end{tabular}
\end{table}

\subsubsection{AViT ablation: noise injection and rollout}
\label{apdx:active_matter_avit_ablation}

\begin{figure}
    \centering
    \includegraphics[width=1.0\linewidth]{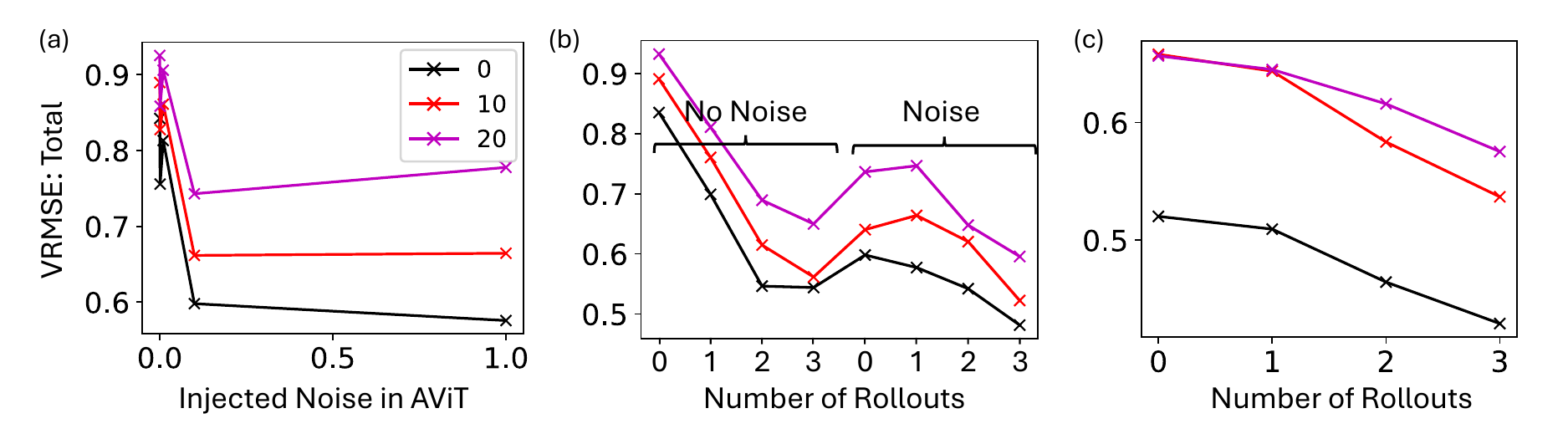}
    \caption{Hyperparameter optimization for dynamics training.
    (a) optimization of the standard deviation of the injected noise into the dynamics model's input. (b,c) optimization of the number of fine-tuning rollout steps.
    (b) contrasts noise injection based pretraining of the dynamics model against pretraining without noise.
    (c) reports performance with a autoencoder trained with hamming noise injection.
    Rollout fine tuning monotonically improves all three.}
    \label{fig:avit_ablation}
\end{figure}

We also optimized the training strategy on the AViT model.
As discussed in the main body of this article, we explored two optimizations.
Figure~\ref{fig:avit_ablation}a illustrates that injecting noise into the AViT's input during pretraining produces a significant improvement in performance, present across all starting times. 
Notably, this optimization is sensitive to the amount of noise.
Additionally, when the amount of noise nears the width of the latent distribution (1.0), performance begins to decay. 

Second, we optimized fine-tuning rollouts.
Figure~\ref{fig:avit_ablation}b,c contrast improvement for the case where pre-training was performed with no noise injection, with noise injection, and with Hamming noise in the autoencoder and noise injection in the AViT.
In all cases, rollout produces significant performance gains with no observable saturation in improvement.
We note that we observed rapid saturation for the spinodal decomposition problem which indicates that the spinodal decomposition problem has saturated as a benchmark.

\subsubsection{Full parameter exploration}
\label{apdx:full_parameter_optimization}

\begin{figure}
    \centering
    \includegraphics[width=0.7\linewidth]{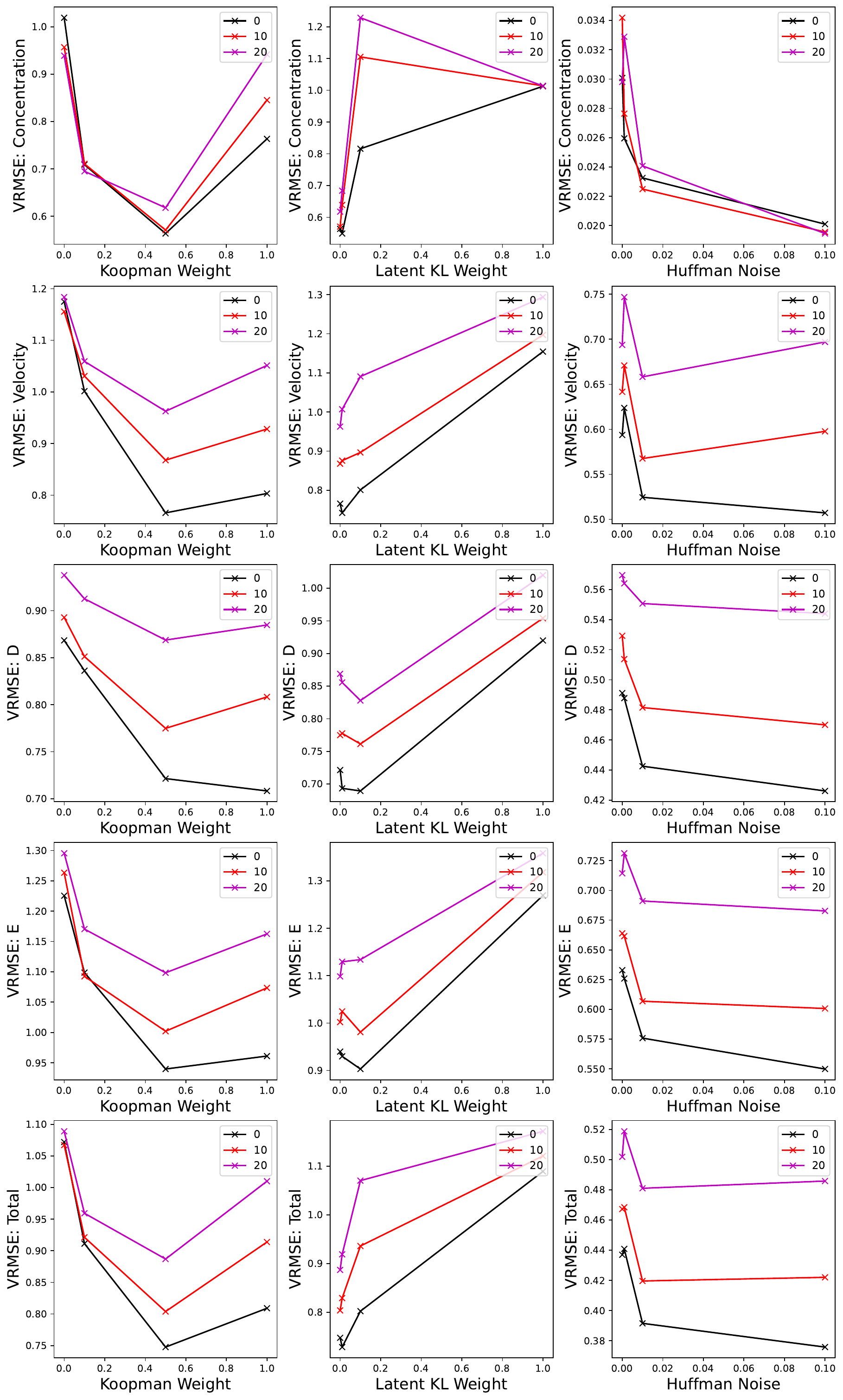}
    \caption{Summary of hyperparameter optimization. Each figure reports $15$-step VRMSE error for its corresponding field variable. Average error over the validation set is reported for 3 starting times: Time Steps $0$, $10$, and $20$. At each time step, the first $9$ steps are provided to the model as context to begin rollout.}
    \label{fig:key_parameter_optimization_complete}
\end{figure}

Figure~\ref{fig:key_parameter_optimization_complete} summarizes the optimization of the Koopman weight, Latent KL weight, and Hamming noise training additions. Both the Hamming and Koopman additions produce consistent improvements across all input parameters. In contrast, the KL divergence is mixed, with significant decreases in performance accompanying most variables. 

\subsubsection{Accumulated improvement: RelMSE}

Table~\ref{tab:relmse_by_dt} reports the breakdown of error improvement using RelMSE as an error. 

\begin{table}[ht]
\centering
\caption{
\textbf{Cumulative ablation of training-level interventions.}
Error by model intervention and rollout length $\Delta t$ on
the active matter benchmark; each row adds one intervention cumulatively on top of all rows above it,
in the order introduced in Section~\ref{sec:interventions}. \emph{Relative} reports the percent change
from the row directly above; \emph{Absolute} reports the percent change from the baseline (row 1).}
\label{tab:relmse_by_dt}
\begin{tabular}{|lccc|ccc|ccc|}
\hline
Model & \multicolumn{3}{c|}{$\Delta t=1$} & \multicolumn{3}{c|}{$\Delta t=5$} & \multicolumn{3}{c|}{$\Delta t=15$} \\
\cline{2-10}& Error & Relative & Absolute & Error & Relative & Absolute & Error & Relative & Absolute \\
\hline
Baseline & 0.07137 & -- & -- & 0.40241 & -- & -- & 0.95216 & -- & -- \\
Spatial & 0.00152 & 0.97871 & 0.97871 & 0.06127 & 0.84775 & 0.84775 & 0.85100 & 0.10625 & 0.10625 \\
Koopman & 0.00130 & 0.14157 & 0.98172 & 0.08094 & -0.32116 & 0.79885 & 0.59237 & 0.30391 & 0.37786 \\
KL & 0.00137 & -0.05239 & 0.98076 & 0.07923 & 0.02119 & 0.80311 & 0.58977 & 0.00440 & 0.38060 \\
Hamming noise & 0.00125 & 0.09187 & 0.98253 & 0.08019 & -0.01212 & 0.80073 & 0.61107 & -0.03612 & 0.35823 \\
Dynamic Noise & 0.00172 & -0.37803 & 0.97593 & 0.05745 & 0.28354 & 0.85723 & 0.44645 & 0.26940 & 0.53112 \\
Roll 1 & 0.00364 & -1.11652 & 0.94905 & 0.04738 & 0.17523 & 0.88225 & 0.43781 & 0.01935 & 0.54020 \\
Roll 2 & 0.00707 & -0.94394 & 0.90095 & 0.03704 & 0.21839 & 0.90796 & 0.38847 & 0.11269 & 0.59201 \\
Roll 3 & 0.00824 & -0.16523 & 0.88459 & 0.03429 & 0.07404 & 0.91478 & 0.34109 & 0.12196 & 0.64177 \\
\hline
\end{tabular}
\end{table}

\subsection{Crystal plasticity cyclic fatigue}\label{appdx:fatigue-expt}

\begin{figure}
    \centering
    \includegraphics[width=1.0\linewidth]{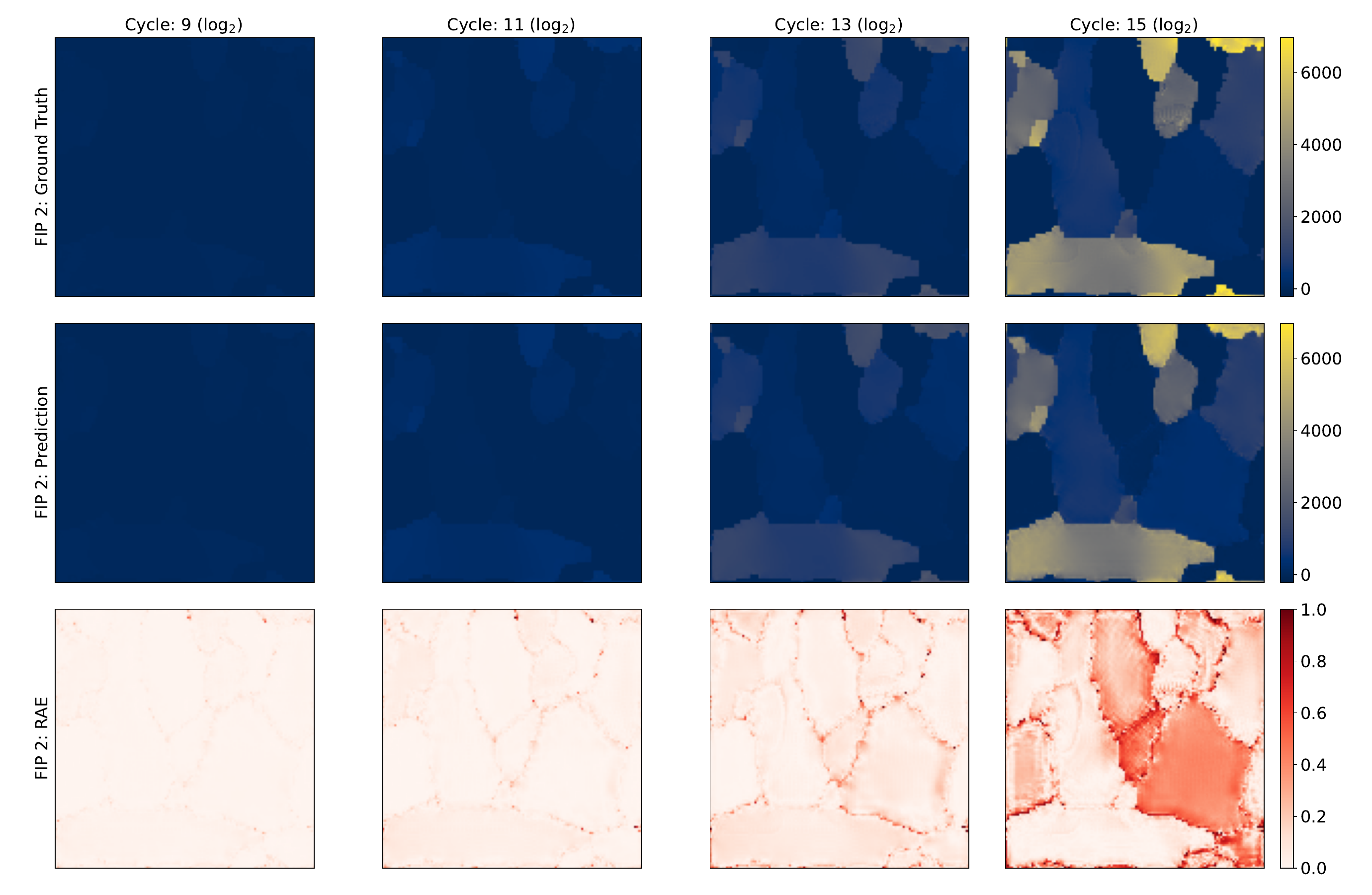}
    \caption{\textbf{Comparison of the MASSIF Ground Truth FIP 2 field against the prediction from the LDM (FIP) model for long rollout example 1.} Row 3 reports relative absolute error where each pixel's value is used to define the relative baseline.}
    \label{fig:fatigue_examplerollout_prelim02}
\end{figure}

Figure~\ref{fig:fatigue_examplerollout_prelim02} illustrates example rollout for the first long rollout example considered in the main body, see Figure~\ref{fig:fatigue_overview}.
The visualization corroborates the discussion in the main body: the model is able to accurately make predictions even far outside the time scales present in its training data. 

\begin{figure}
    \centering
    \includegraphics[width=1.0\linewidth]{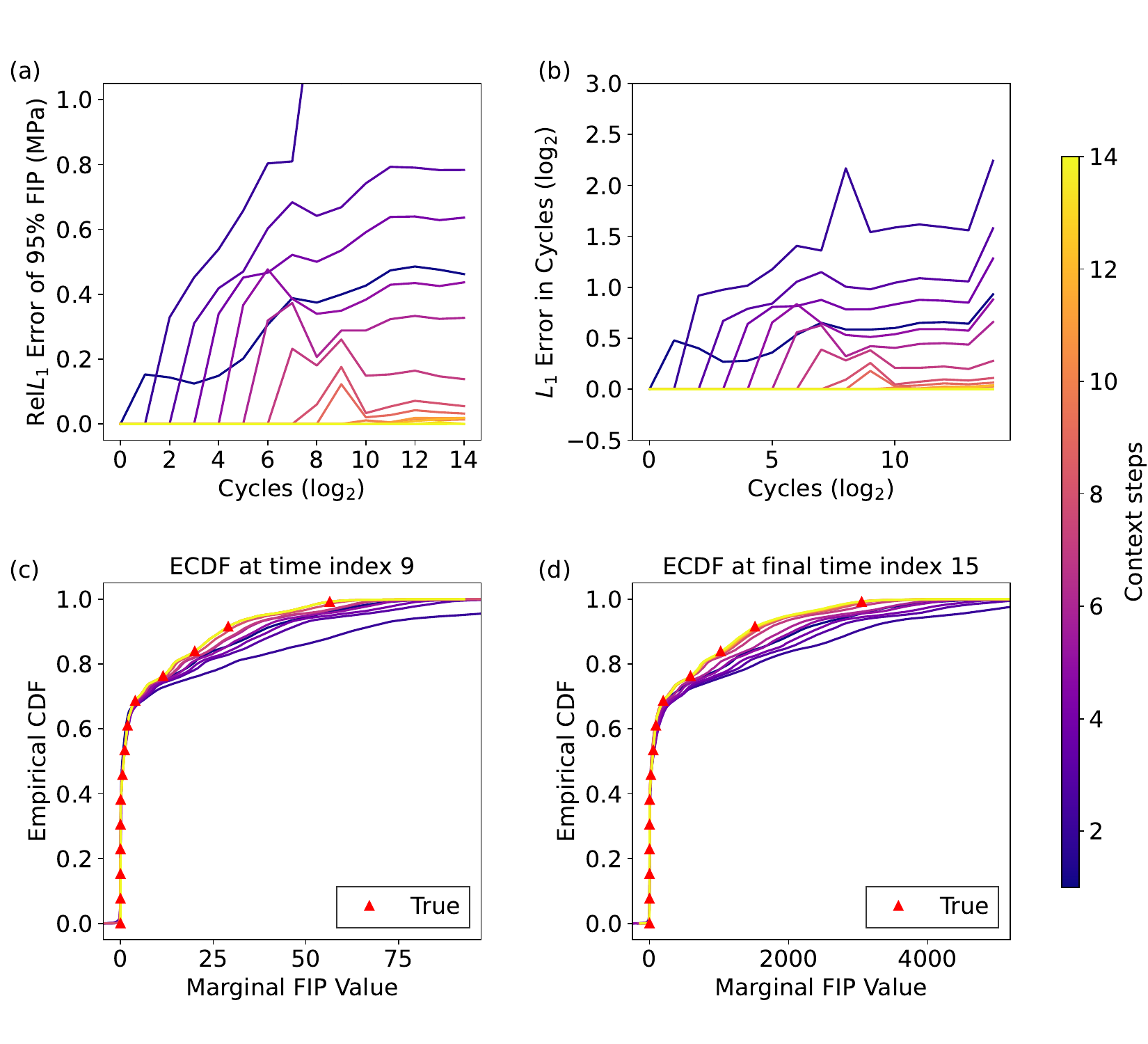}
    \caption{\textbf{Summary of cyclic fatigue based failure prediction on long rollout Example 2 using the LDM (FIP) model.} In all plots, lines are colored by the number of simulation context steps provided before switching to the neural solver. (a) the relative $L_1$ error of the predicted $95$-percentile FIP at each cycle order of magnitude. (b) A reinterpretation of (a): $L_1$ error in the number of cycles to failure if failure occurred at each cycle order of magnitude. (c) predicted Emperical FIP CDF at the temporal boundary of training data. (d) predicted Empirical FIP CDF at maximum ground truth cycle count ($2^{15}$ cycles).}
    \label{fig:fatigue_fip_prediction_prelim00}
\end{figure}

\begin{figure}
    \centering
    \includegraphics[width=1.0\linewidth]{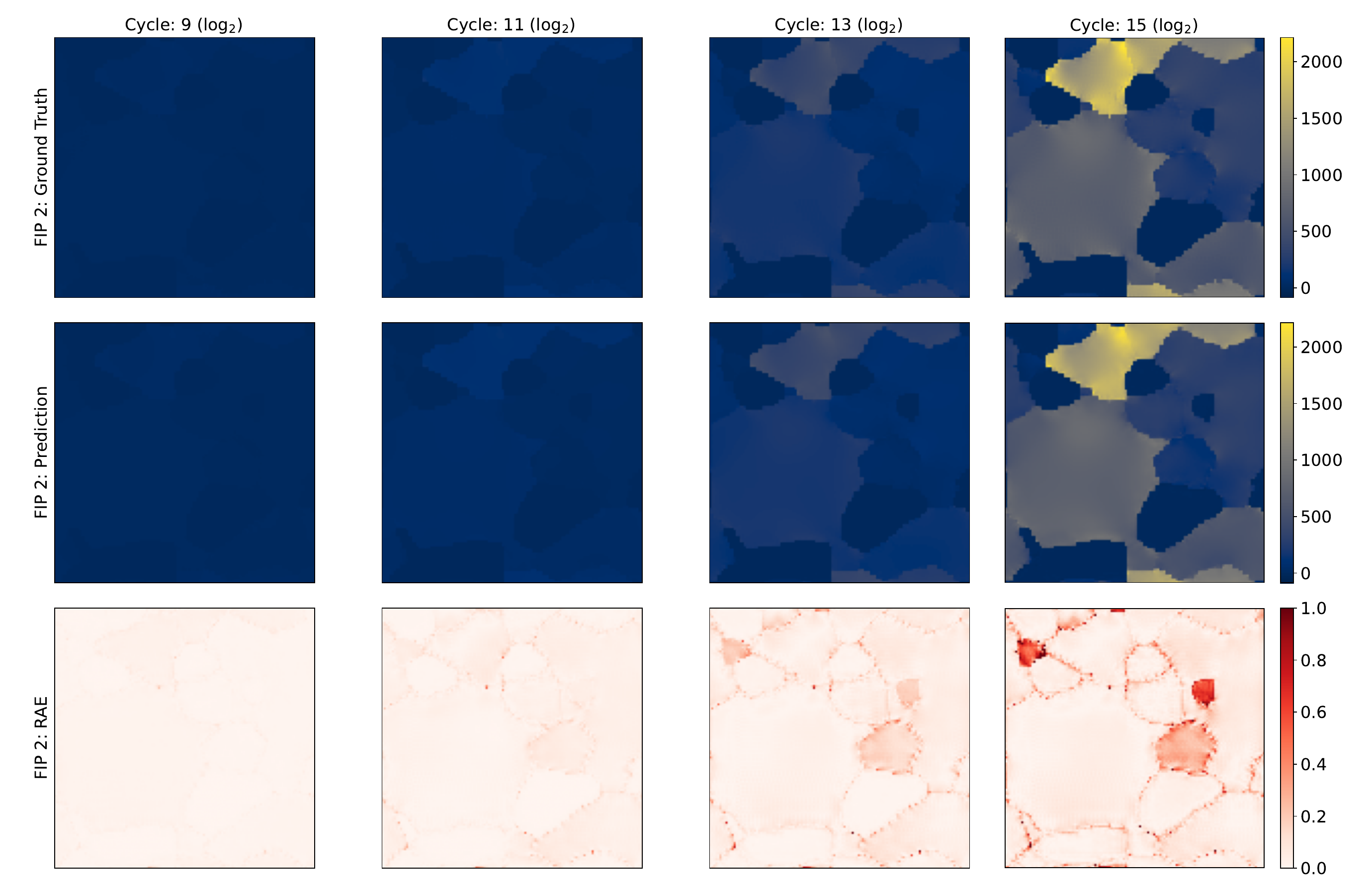}
    \caption{\textbf{Comparison of the MASSIF Ground Truth FIP 2 field against the prediction from the LDM (FIP) model for long rollout example 2.} Row 3 reports relative absolute error where each pixel's value is used to define the relative baseline.}
    \label{fig:fatigue_examplerollout_prelim00}
\end{figure}

Figure~\ref{fig:fatigue_examplerollout_prelim00} and Figure~\ref{fig:fatigue_fip_prediction_prelim00} summarize performance on the second long rollout example.
This study experienced higher loading conditions which increased the difficulty of prediction and extrapolation.
However, even in this case, conditioned on $2^5$ cycles of simulation results, the model is still able to predict the cycles to failure at $2^{15}$ cycles within an order of magnitude.
In this example, we see that the addition of more simulation cycles enables the model to achieve near perfect precision at the $2^{15}$ mark, see Figure~\ref{fig:fatigue_fip_prediction_prelim00}c,d.

\subsubsection{AViT ablation trends}
\label{app:fatigue_avit_ablation_trends}

\begin{figure}
    \centering
    \includegraphics[width=1.0\linewidth]{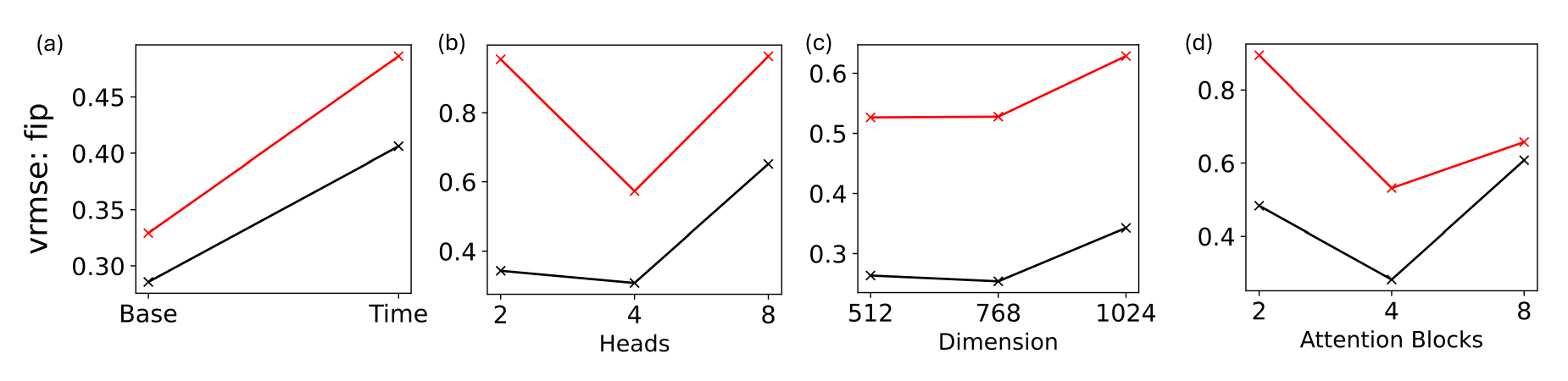}
    \caption{Hyperparameter optimization of the AViT Dynamics model for the Cyclic Fatigue Case Study. (a) comparison of standard training against training where the loss is weighted towards larger times. (b) hyperparameter optimization of the number of heads in the attention mechanism. (c) hyperparameter optimization of the dimensionality of the AViT embedding space. (d) hyperparameter optimization of the number of attention blocks.}
    \label{fig:fatigue_avit_ablation}
\end{figure}

Unlike the active matter study, in this cycle fatigue study we perform minor architecture optimization on the AViT.
Figure~\ref{fig:fatigue_avit_ablation}b,c,d summarizes the main results of optimizing the AViT's number of heads, latent dimension, and number of attention blocks.
We select intermediate values of all for the final models (heads: 4, dimension: 768, blocks: 4). 

In addition, we explored reweighting the loss to promote increased performance for larger times. 
However, we observe that this actually led to a decrease in overall rollout performance, Figure~\ref{fig:fatigue_avit_ablation}a.

\subsubsection{Fatigue cycle estimate}

In the fatigue case study, the model's performance was first quantified by analyzing the error in the $95^{\mathrm{th}}$-percentile FIP prediction.
Although this value is the most directly calculable relevant quantity, the meaning of its magnitude can be challenging to directly interpret.
As a result, in Fig.~\ref{fig:fatigue_fip_predictions}b we present the error in the predicted cycles.
This prediction is derived from the $95^{\mathrm{th}}$-percentile prediction via a Taylor series analysis. 

Specifically, for each time step, we have a calculated true $95^{\mathrm{th}}$-percentile, $p_T$, (extracted from the DNS calculations), a predicted $95^{\mathrm{th}}$-percentile, $p_*$, and a true time $s_T$.
Via Taylor series expansion of the $95^{\mathrm{th}}$-percentile function, we can write the following expression for the predicted time step:

\begin{equation}
    s_* = (p_* - p_T) \left. \left(\frac{dp}{ds} \right)^{-1}  \right|_{s=s_T} + s_T
\end{equation}

\noindent We estimate the gradient term using second order finite differences.
This expression is used and plotted in Fig.~\ref{fig:fatigue_fip_predictions}b.

\newpage
\bibliographystyle{ieeetr}  
\bibliography{references} 

\end{document}